\documentclass[10pt,twocolumn,letterpaper]{article}

\usepackage[pagenumbers]{cvpr} 

\usepackage[dvipsnames]{xcolor}

\definecolor{cvprblue}{rgb}{0.21,0.49,0.74}
\usepackage[pagebackref,breaklinks,colorlinks,citecolor=cvprblue]{hyperref}

\usepackage[ruled]{algorithm2e}
\usepackage{bbm}
\usepackage{bm}
\usepackage{color}
\usepackage{multirow}
\usepackage{array}
\usepackage{booktabs}
\usepackage{diagbox}
\usepackage{bbding}
\usepackage{colortbl}
\usepackage{tabularx}

\newcommand{\Tref}[1]{Table~\ref{#1}}

\newcommand{\Fref}[1]{Figure~\ref{#1}}
\newcommand{\Sref}[1]{Section~\ref{#1}}

\newcommand{\eref}[1]{Eq.~(\ref{#1})}

\newcommand{\sref}[1]{Sec.~\ref{#1}}

\def\paperID{3940} 
\def\confName{CVPR}
\def\confYear{2024}

\title{OPERA: Alleviating Hallucination in Multi-Modal Large Language Models \\ via Over-Trust Penalty and Retrospection-Allocation }

\author{
Qidong Huang\textsuperscript{\rm 1,2,}\thanks{Work done during an internship in Shanghai AI Laboratory.},
Xiaoyi Dong\textsuperscript{\rm 2,3},
Pan Zhang\textsuperscript{\rm 2},
Bin Wang\textsuperscript{\rm 2},
Conghui He\textsuperscript{\rm 2},
Jiaqi Wang\textsuperscript{\rm 2},\\
Dahua Lin\textsuperscript{\rm 2},
Weiming Zhang\textsuperscript{\rm 1},
Nenghai Yu\textsuperscript{\rm 1}\\
\textsuperscript{\rm 1}Anhui Province Key Laboratory of Digital Security, University of Science and Technology of China \\
\textsuperscript{\rm 2}Shanghai AI Laboratory \quad \
\textsuperscript{\rm 3}The Chinese University of Hong Kong \\
{\tt\small \{hqd0037@mail., zhangwm@, ynh@\}ustc.edu.cn}\quad
{\tt\small \{xydong@, dhlin@\}ie.cuhk.edu.hk} \\
{\tt\small\{zhangpan@, wangbin@, heconghui@\}pjlab.org.cn} \quad
{\tt\small wjqdev@gmail.com}
}

\begin{document}
\maketitle

\begin{abstract}
Hallucination, posed as a pervasive challenge of multi-modal large language models (MLLMs), has significantly impeded their real-world usage that demands precise judgment.
Existing methods mitigate this issue with either training with specific designed data or inferencing with external knowledge from other sources, incurring inevitable additional costs. 
In this paper, we present \textbf{OPERA}, a novel MLLM decoding method grounded in an \textbf{O}ver-trust \textbf{Pe}nalty and a \textbf{R}etrospection-\textbf{A}llocation strategy, serving as a nearly \textbf{free lunch} to alleviate the hallucination issue without additional data, knowledge, or training. 
Our approach begins with an interesting observation that, most hallucinations are closely tied to the knowledge aggregation patterns manifested in the self-attention matrix, \ie, 
MLLMs tend to generate new tokens by focusing on a few summary tokens, but not all the previous tokens. 
Such partial over-trust inclination results in the neglecting of image tokens and describes the image content with hallucination.
Based on the observation, OPERA introduces a penalty term on the model logits during the beam-search decoding to mitigate the over-trust issue, along with a rollback strategy that retrospects the presence of summary tokens in the previously generated tokens, and re-allocate the token selection if necessary. 
With extensive experiments, OPERA shows significant hallucination-mitigating performance on different MLLMs and metrics, proving its effectiveness and generality.
Our code is available at: \href{https://github.com/shikiw/OPERA}{This link}.

\end{abstract}

\section{Introduction}
\label{sec:intro}

\begin{figure}[t]
\centering
\includegraphics[width=1.0\linewidth]{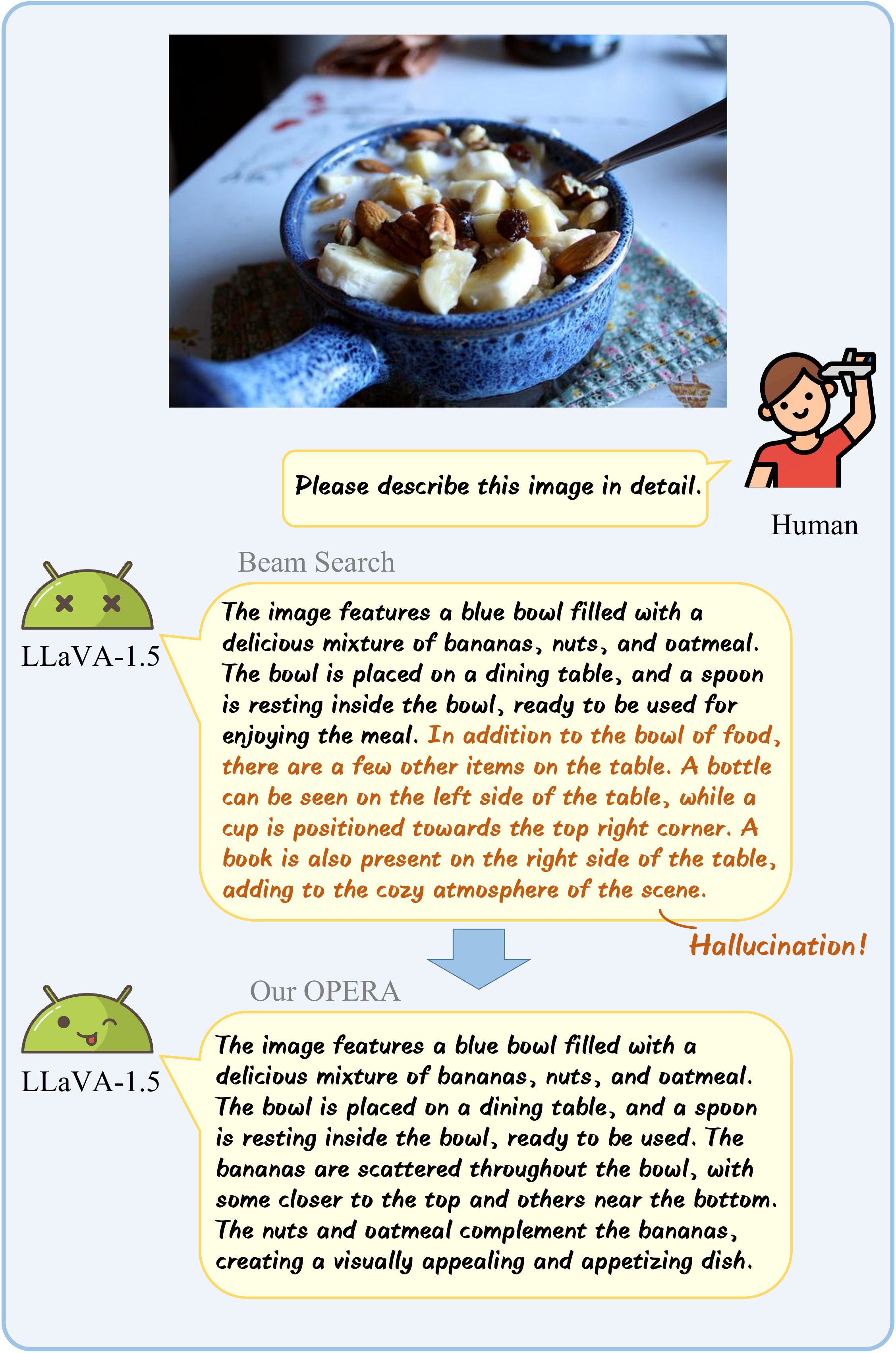}
\caption{OPERA's performance on reducing hallucinations. }
\label{fig:teaser}
\end{figure}

Recent advancements in multi-modal large language models (MLLMs) \cite{liu2023llava,zhu2023minigpt,instructblip,bai2023qwen,zhang2023internlm,chen2023shikra,liu2023improvedllava,dong2024internlm} has greatly elevated general-purpose foundation models to unprecedented levels. 
These models enable users to interact using images as input, facilitating free-flowing communication based on the content of these images. 
The impressive abilities of MLLM allows it to be adept at a variety of vision tasks \cite{zhang2023gpt4roi,li2023llavamed,black2023training}, meanwhile easily handling some complex content comprehension \cite{lai2023lisa} or generation \cite{brooks2023instructpix2pix,geng2023instructdiffusion}. 

Notwithstanding their remarkable versatility, MLLMs also grapple with a significant challenge known as the ``hallucination'' problem.
Specifically, MLLMs often hallucinate incorrect statements to the user-provided image and prompts, \eg, producing irrelevant or nonsensical responses, indentifying inaccurate objects in terms of colors, quantities and locations that do not exist in the image. 
This flaw poses substantial risks for practical applications of MLLMs to become a trustworthy assistant. 
For instance, in model-assisted autonomous driving scenarios, such misinterpretations of road scene images may lead to wrong judgments of system and serious traffic accidents. 

Various approaches \cite{liu2023aligning,zhou2023analyzing,yin2023woodpecker,wang2023vigc} have been proposed to reduce hallucinations in MLLMs. While these methods incur substantial additional costs, including the annotation budget for extra instruction data for training \cite{liu2023aligning}, the integration of external knowledge or models, \etc.

\begin{figure}[t]
\centering
\includegraphics[width=1\linewidth]{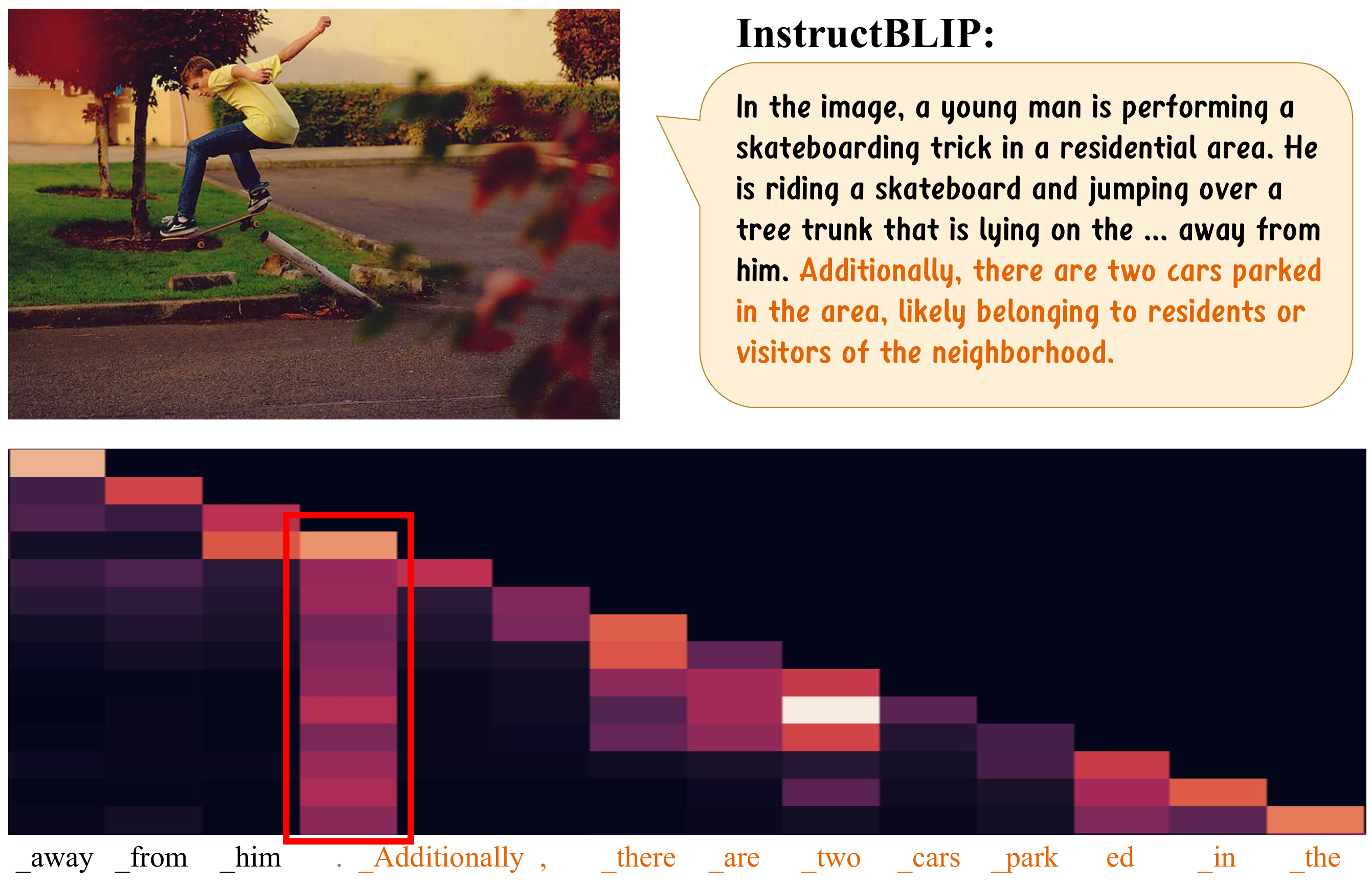}
\caption{A case of relationship between hallucinations and knowledge aggregation patterns. Hallucinations are highlighted.}
\label{fig:intro2attn}
\end{figure}

\begin{figure}[t]
\centering
\includegraphics[width=1\linewidth]
{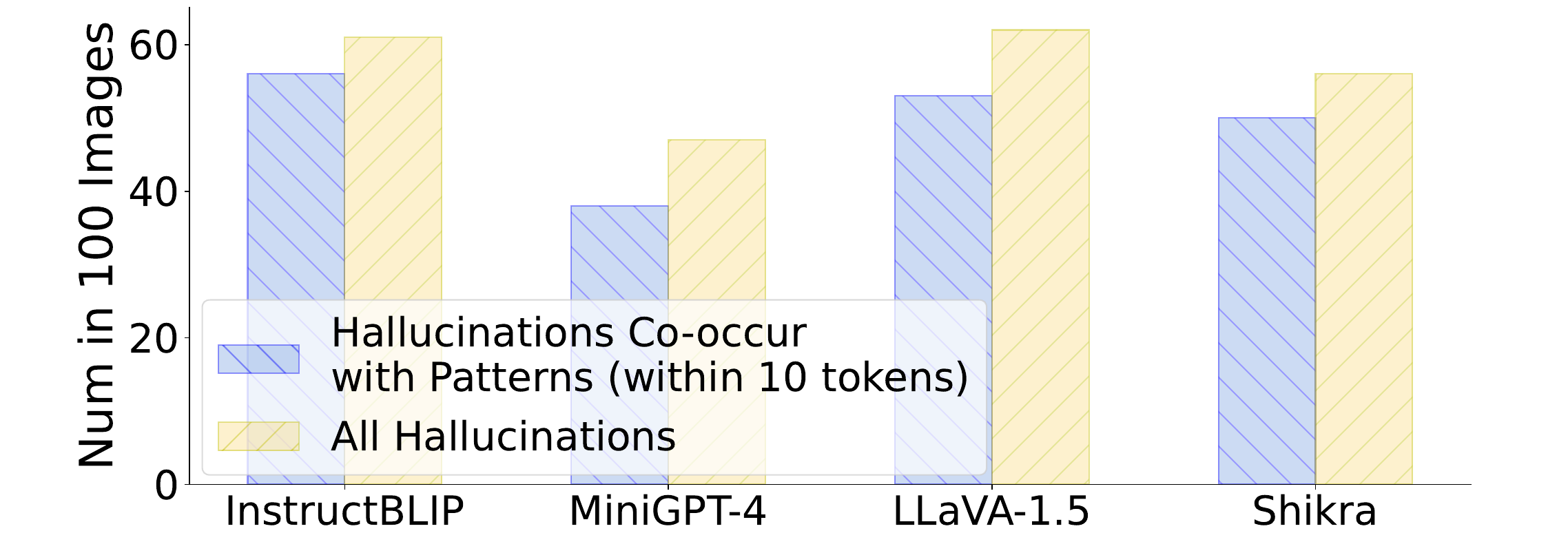}
\caption{Hallucinations often start within the first 10 tokens after knowledge aggregation patterns.}
\label{fig:intro2hist}
\end{figure}

In this paper, we delve into the challenge of mitigating MLLMs' hallucination during inference, without introducing additional data, models, or knowledge. 
Our investigation commences with a noteworthy `\textbf{partial over-trust}' observation found while visualizing self-attention maps for decoded sequences. 
As illustrated in \Fref{fig:intro2attn}, we discern a recurring pattern where the inception of many hallucinated contents aligns with the subsequent tokens generated after a columnar attention pattern. 
Notably, these columnar attention patterns often manifest on tokens that lack substantial informativeness, \eg, full stop or quotation marks. 
Intuitively, this peculiarity reveals a weird fact that, a token exhibiting a columnar attention pattern typically possesses limited information, yet exerts a pronounced influence on the prediction of all subsequent tokens. 
Moreover, as shown in \Fref{fig:intro2hist}, we find that most of the subsequent contents contain reasoning or hallucinations.

\begin{figure}[t]
\begin{minipage}{0.31\linewidth}
    \centering
    \includegraphics[width=1\linewidth]{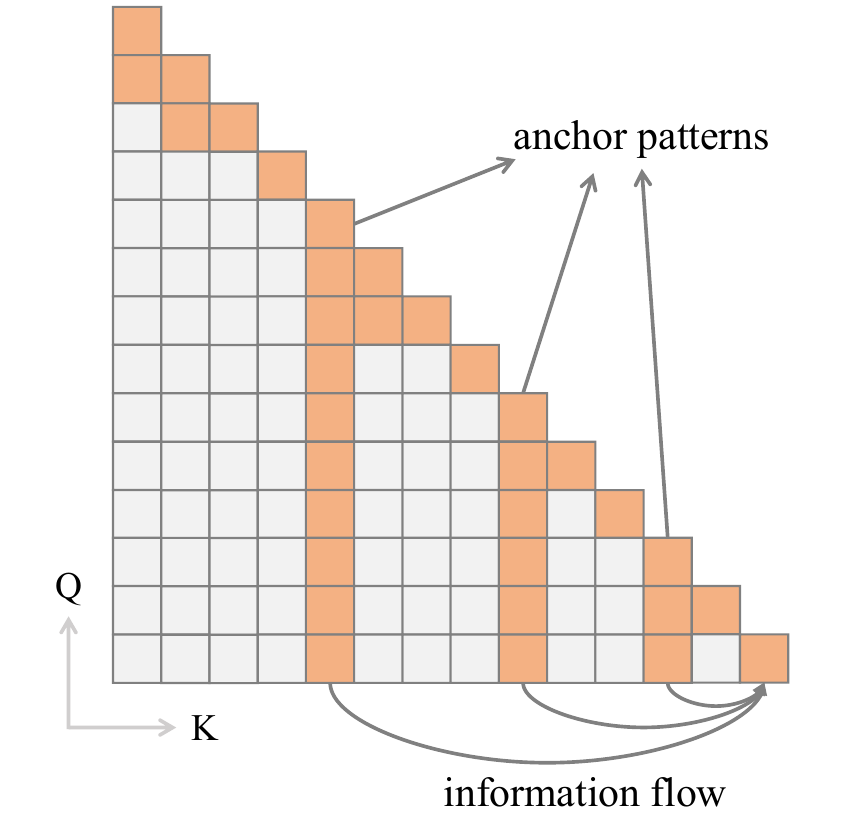}
    \\
    \scriptsize (a)
\end{minipage}
\hfill
\begin{minipage}{0.33\linewidth}
    \centering
    \includegraphics[width=1\linewidth]{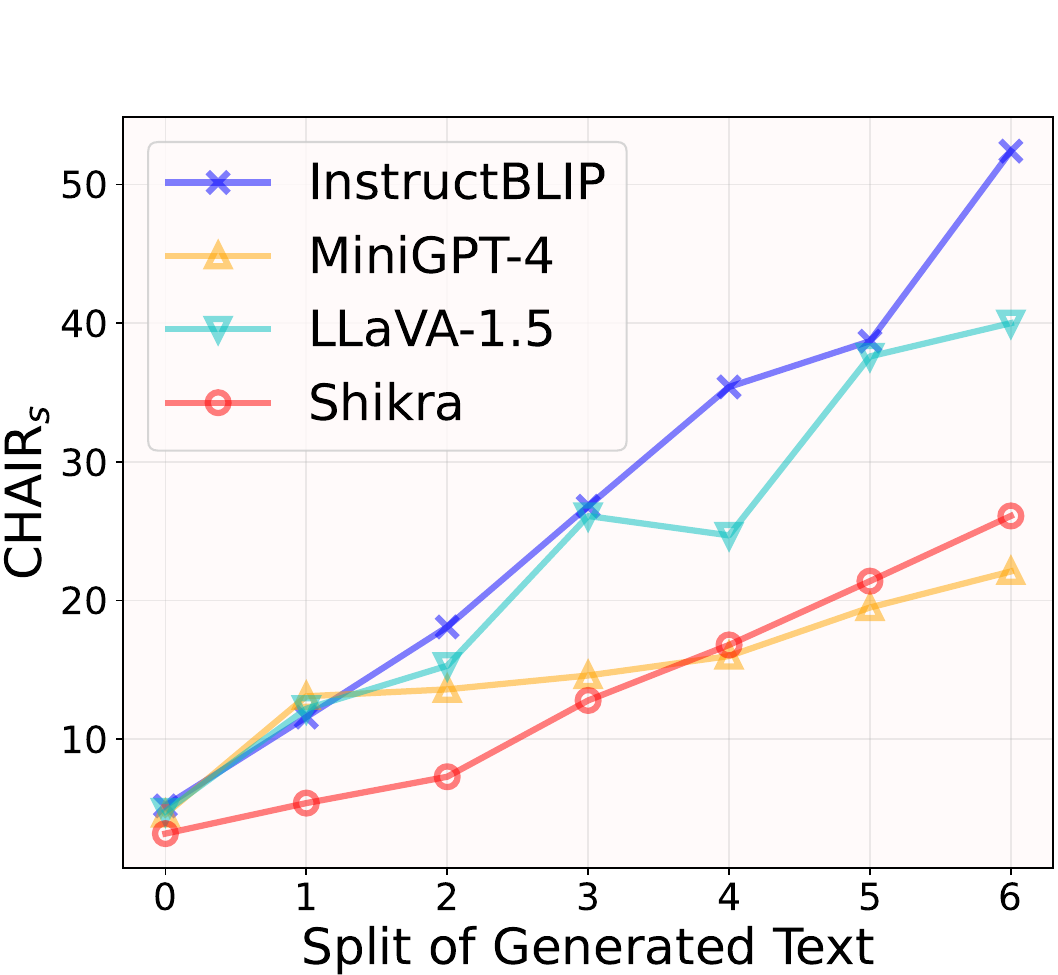}
    \\
    \scriptsize (b)
\end{minipage}
\begin{minipage}{0.33\linewidth}
    \centering
    \includegraphics[width=1\linewidth]{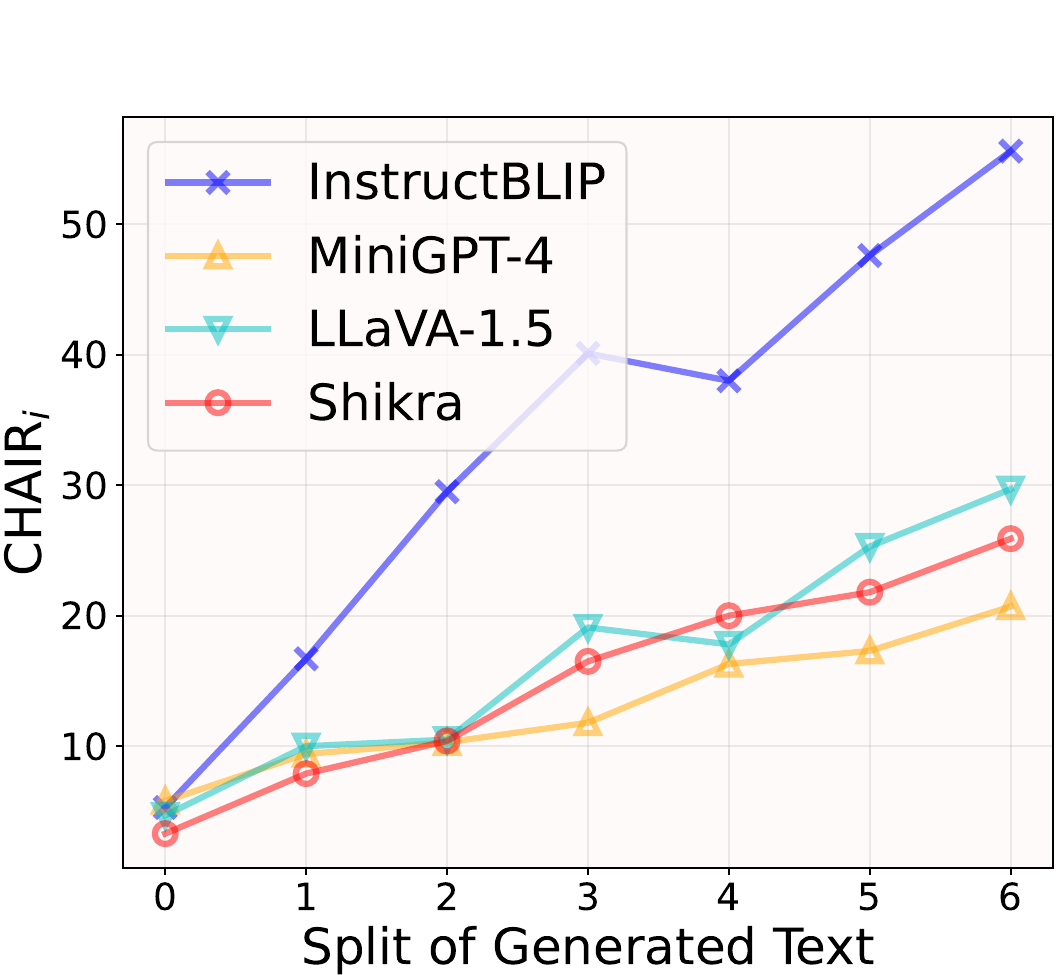}
    \\
    \scriptsize (c)
\end{minipage}
\caption{(a) The aggregation pattern is consistent with recent `anchor token' observation. (b), (c) show the increasing CHAIR scores (more hallucinations) on 5,000 randomly selected MSCOCO images when more anchor tokens appear in the context.}
\label{fig:evidence}
\end{figure}

\textbf{`Aggregation pattern' seems to be the nature of LLM.} 
We hypothesize that such tokens serve as summary tokens, which aggregate the crucial knowledge from previous tokens in the sequence and guide the subsequent tokens generation. 
Our observation is consistent with the recent `anchor token' \cite{wang2023label} observation in the NLP area, which finds the LLM tends to aggregate previous information on a few anchor tokens at shallow layers and predict the next token based on these anchors at the deep layer (\Fref{fig:evidence}(a)). 

\textbf{`Aggregation pattern' leads to hallucination of current MLLMs}. 
Current MLLMs usually put the vision tokens at the beginning of the sequence, and they are expected to focus on the vision tokens and provide an precise understanding. 
However, as the generated text goes longer, it will be easier for vision information to be attenuated during the transmission of information between summary tokens (a single summary token can not remember the dense and rich information given by the whole context).
In detail, the subsequent tokens may ignore the forehead image tokens and over-trust the closer summary tokens via their stronger attention attended, leading to hallucinations raised by the model bias, \eg, hallucinating ``cars'' based on the ``road'' mentioned in the previous sentence. 
In other words, the more summary tokens appear, the more easily MLLM hallucinations are induced. 
To prove it, we split the long responses of MLLMs based on the position of summary tokens, and calculate the CHAIR scores for different splits separately. 
As shown in \Fref{fig:evidence}(b)(c), the CHAIR score shows a clear \textbf{positive relation} with the split number of the generated text, \ie, more hallucinations are generated when more summary tokens appear in the context, manifested as the co-occurrence of them.

To alleviate the partial over-trust issue, we present OPERA, a novel MLLM decoding approach grounded in an \textbf{O}ver-trust \textbf{Pe}nalty and a \textbf{R}etrospection-\textbf{A}llocation strategy. 
The over-trust penalty introduces a weighted score for the candidate selection step in the Beam Search \cite{graves2012sequence,sutskever2014sequence,boulanger2013audio}, so that the candidate with an over-trust pattern will have lower priority to be selected.
Specifically, for each decoding token, we investigate the local window segmented on the self-attention map of the decoded sequence, and devise a column-wise metric to calculate the intensity of knowledge aggregation patterns. 
This metric produces a value that indicates the over-trust degree between in-window tokens and the summary tokens. 
It is naturally incorporated with the model logits predicted for the next token in the Beam Search and penalizes the appearance of over-trust patterns. 
Further, considering the hysteresis of the appearance of the knowledge aggregation pattern,  the hallucination may exist in all the candidates when it can be observed. We propose a retrospection-reallocation strategy to help the decoding process roll back to the position of the summary token and re-select better candidates that can avoid such a pattern.  
Such retrospection is triggered when the location overlap of the maximum of in-window penalty scores reaches a threshold. 

With extensive experiments on benchmarks and hallucination metrics, along with GPT-4/GPT-4V assessments, OPERA demonstrates the generalized hallucinations-reducing performance on various MLLM models. 
Our contributions can be summarized as follows:
\begin{itemize}
    \item Our OPERA alleviates the MLLMs' hallucination issue during inference, without introducing any external data, knowledge, or additional training.
    \item We reveals the appearance of hallucinations and over-trust patterns, and propose a penalty-based decoding method equipped with retrospection-reallocation strategy. 
    \item Extensive evaluation including GPT assessments prove the superior performance of OPERA, which serves as a nearly free-lunch to mitigate hallucinations.
\end{itemize}

\section{Related Work}
\label{sec:related_work}

\subsection{Multi-Modal Large Foundation Models}

Recent progresses of computational resources has greatly facilitated the research into large-scale foundational models incorporated with multi-modal learning. 
Powered by open-sourcing large language models such as LLaMA \cite{touvron2023llama,touvron2023llama2} and Vicuna \cite{chiang2023vicuna}, 
MLLMs \cite{liu2023llava,zhu2023minigpt,instructblip,huang2021initiative,bai2023qwen,huang2023diversity,huang2022shape,chen2023sharegpt4v} understand and generate diverse content in a more comprehensive way by integrating information from different modalities, such as text, images, and audio. 
The series of CLIP and BLIP well aligns the text features and image features. 
LLaVA \cite{liu2023llava}, InstructBLIP \cite{instructblip} and MiniGPT-4 \cite{zhu2023minigpt} take a step forward in this field, allowing users to interact with these intelligence with images and texts as prompts. 
All of them share the same two training phases, \ie, pre-trained feature alignment and instruction fine-tuning, to help the model to comprehend the format of instruction input. 
Shikra \cite{chen2023shikra} incorporates grounding data and teaches the model to understand the grounding knowledge in the given images. 
All of aforementioned MLLM models suffer from severe hallucination problems. 
Consequently, we mainly conduct the experiments on these four models in our paper.

\subsection{Hallucination in Large Foundation Models}

The hallucination \cite{ji2023survey,zhang2023language} refers to the generation of text that is either irrelevant, factually incorrect, or nonsensical in the given context, which is quite severe in current large foundation models. 
This issue can arise due to overfitting to specific patterns in the training data, lack of understanding of real-world facts, or an inability to effectively contextualize the given input.
The primary concern regarding hallucination in LLMs is the factual accuracy of generated content, \ie, conflicting with world knowledge or common sense. 
In MLLMs, the primary worry centers around faithfulness, \ie, assessing whether the generated answers conflict with user-provided images.
Researches on mitigating current LLMs' hallucination issues often focuses on several aspects, including refining the training process, using larger and more diverse datasets \cite{lee2022factuality}, or implementing post-training evaluation \cite{du2023improving} and correction mechanisms \cite{peng2023check,manakul2023selfcheckgpt}. 
While for MLLMs, relevant researches are still quite few \cite{liu2023aligning,zhou2023analyzing,yin2023woodpecker}. 
However, most of these countermeasures have a large drawback that, they either introduce large quantities of extra data, or resort to more powerful external models or knowledge. 
Compared with them, our OPERA serves as nearly free lunch for alleviating the hallucination issue, which does not incur extra training, data, or knowledge.

\subsection{Decoding Strategy in Language Models}
Decoding strategies in language models are crucial for determining how these models generate text. They play a pivotal role in shaping the output's quality, relevance, and coherence. 
Greedy Decoding simply selects the most likely next word at each step. 
While fast and computationally efficient, greedy decoding often leads to repetitive and less varied text.
Beam Search \cite{graves2012sequence,sutskever2014sequence,boulanger2013audio} is a more sophisticated approach, beam search keeps track of a predefined number of hypotheses at each step, expanding on them to find a more optimal sequence. 
Top-k Sampling \cite{fan2018hierarchical} adds randomness to the generation process by randomly selecting from the top-k likely next words, introducing diversity in the output but can sometimes produce less coherent results.
Top-p (Nucleus) Sampling \cite{holtzman2019curious} is an evolution of Top-k, Nucleus sampling considers a dynamic number of words that cumulatively reach the probability $p$. This method provides a balance between randomness and relevance, often leading to more coherent and interesting outputs than Top-k sampling.
DoLa \cite{chuang2023dola} decoding is a recently proposed decoding method that aims to mitigate the hallucinations in MLLMs, which contrasts the logits of mature layer and pre-mature layers and rescale the increments as the output. 
In this paper, we compare our proposed OPERA with these common decoding strategies, focusing on the performance on the hallucination issues of MLLMs.

\section{Method}
\label{sec:method}

In the following, we first formulate the generation procedure of the MLLMs for the easy understanding of our OPERA, then introduce the calculation of the proposed Over-Trust Logit Penalty and Retrospection-Allocation Strategy respectively.

\subsection{Formulation of MLLMs Generation}
\label{mllm_gen}
The generation procedure of LLMs could be parsed into three components: input formulation, model forward, decoding.

\noindent\textbf{Input Formulation}.  
The input of MLLMs contains both image and text. Putting aside the specific architecture difference, the MLLMs commonly use a vision encoder to extract visual tokens from the raw images, and map them into the LLMs' input space with a cross-modality mapping module. The mapped visual tokens are used as part of the LLM input, along with the text input. We denote the visual tokens as $\mathbf{x}^v = \{x_0,x_1,\ldots,x_{N-1}\}$. Here $N$ is the length of the visual tokens and it is a fixed number in most cases.  
Correspondingly, the input text is tokenized with the tokenizer and we denote it as $\mathbf{x}^p = \{x_N,x_{N+1},\ldots,x_{M+N-1}\}$.
The image and text tokens are concatenated as the final input sequence and we denote it as  $\{x_i\}_{t=0}^{T-1}$ that $T=N+M$.

\noindent\textbf{Model Forward}.  
The MLLM is trained in an auto-regressive manner with a causal attention mask, each token predicts its next token based on previous tokens, formally:
\begin{equation}
\begin{aligned}
    \mathbf{h} &= \mathrm{MLLM}(\mathbf{x}_i) \\ 
    \mathbf{h} &= \{h_0,h_1,\ldots,h_{T-1}\}
\end{aligned}
\end{equation}
where $\mathbf{h}$ is the output hidden states of the last layer of the MLLM.

Next, MLLMs use a vocabulary head $\mathcal{H}$ to project the hidden states $\mathbf{h}$ and get the logits (or probabilities) for the next token prediction, formally:
\begin{equation}
    p(x_t|x_{<t}) = \text{SoftMax}[\mathcal{H}(h_t)]_{x_t}, \quad x_t\in \mathcal{X},
\end{equation}
where we use $x_{<t}$ to simplify the sequence $\{x_i\}_{i=0}^{t-1}$ and $\mathcal{X}$ means the whole vocabulary set. 

\noindent\textbf{Decoding}.  
Based on the logits $p(x_t|x_{<t})$, there are several decoding strategy developed, including Greedy Decoding, Beam Search, DoLa, \etc. The decoded token is concatenated to the last of the original input text for the next-round generation, until the generation is ended. 

Our OPERA is based on the Beam Search \cite{graves2012sequence,sutskever2014sequence,boulanger2013audio}, which is a accumulated-score-based decoding strategy. Briefly, With a given beam size $N_{beam}$, the Beam Search keeps $N_{beam}$ candidate sequences, where each candidate is a decoded sequence $\mathbf{x}^{N_{beam}}$ with a beam score. When decoding token $x_t$, each candidate hypothesis will select $N_{beam}$ candidate tokens based on the Top-$N_{beam}$ probabilities in the logits. And finally, the decoding procedure will output the hypothesis wins the best beam score.

\subsection{Over-Trust Logit Penalty}
\label{sec:olp}
As we analyzed in Sec.\ref{sec:intro}, there exists a high-probability co-currence between the hallucination and the knowledge aggregation patterns. However, such pattern has a significant \textbf{hysteresis}, \ie, 
the patterns can not be immediately observed when the corresponding token is decoded, but after several subsequent tokens been decoded, and the hallucination may already occurred. 

In response to the hysteresis, we propose `Over-Trust Logit Penalty', an \textit{accumulative penalty weighted in the beam score}, which influences the selection of both the current token and the candidate sequence. A candidate sequence accumulated with a large penalty will have a lower priority to be selected so that the output with hallucinations will be possibly omitted.

\begin{figure}[t]
\centering
\includegraphics[width=1.0\linewidth]{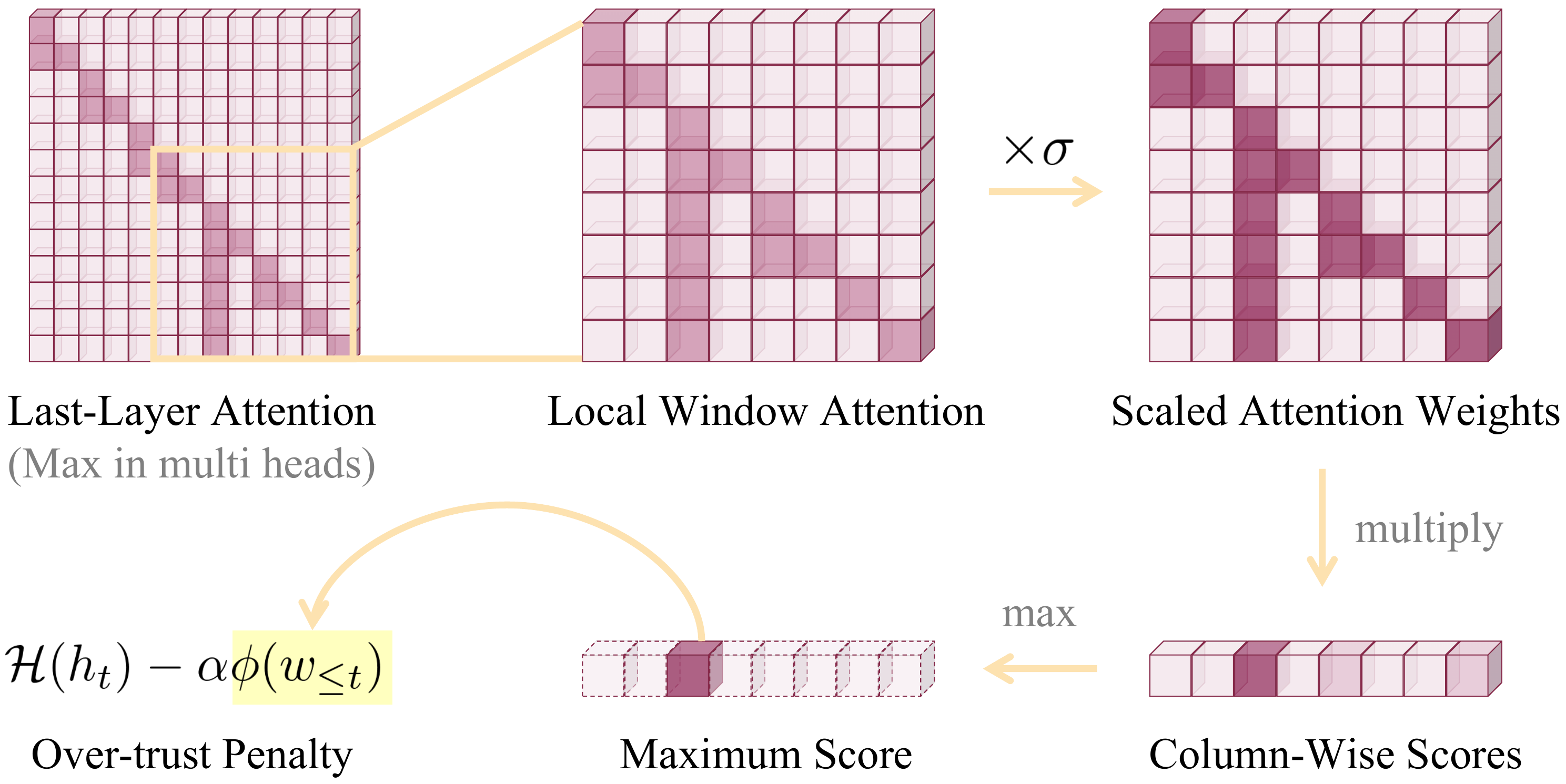}
\caption{The scheme of calculating the proposed over-trust penalty term. We first cut out a local window on the self-attention map, then we scale up the values and conduct the column-wise multiplication to get a score vector, finally we choose the maximum score as the penalty term.}
\label{fig:method1}
\end{figure}

In practice, we investigate a local window on the self-attention weights and leverage column-wise product to calculate the metric values. 
Denote the current generated sequence as $\{x_i\}_{i=0}^{t-1}$ and their casual self-attention weights $\{\omega_{t-1,j}\}_{j=0}^{t-1}$ paid on the next token prediction, in which the weights can be depicted by softmax result as $\omega=\text{SoftMax}(\frac{QK^\top}{\sqrt{D}})$ and $Q$, $K$, $D$ denote query feature, key feature, feature dimension respectively.
We consider to gather all of previous self-attention weights in a local window for characterizing the knowledge pattern, \ie, the local window attention is defined as
\begin{equation}
    \mathbf{W}_{t-1}^k=\{\mathbf{w}^i\}_{i=t-k}^{t-1},\quad\text{s.t. }\mathbf{w}^i=\{\omega_{i,j}\}_{j=t-k}^{i},  
\end{equation}
where $k$ denotes the size of local window we cropped on the attention map, $\omega_{i,j}$ means the attention weight assigned by the $j^{th}$ token to the $i^{th}$ token. 
There are two points should be clarified: 
1) our window does not involve the attention weights of image tokens or prompt tokens because we only concentrate on the knowledge aggregation patterns on generated tokens, \ie, $t-k\geq N+M$. 
2) we select the maximum weight in multi-head attentions and re-normalize the values since it usually indicates the strong confidence of models.

With the local window attention weights $\mathbf{W}_{t-1}^k$, we can calculate upon a simple metric to describe the size of the knowledge aggregation pattern. 
Specifically, we first do some preprocess on $\mathbf{W}_{t-1}^k$, including filling the upper triangle of the matrix with zeros and scaling up the attention values as the values are usually too small, \ie, 
\begin{equation}
    \mathbf{W}_{t-1}^k\triangleq\{\mathbf{w}^i\}_{i=t-k}^{t-1},
    \quad\text{s.t. }\mathbf{w}^i=\{\sigma\omega_{i,j}\}_{j=t-k}^{t-1},  
\end{equation}
where $\{\omega_{i,j}\}_{j=i+1}^{t-1}$ are zeros and $\sigma$ is a configurable scaling factor.

As illustrated in \Fref{fig:method1}, we then conduct the column-wise multiplication on the lower triangle of the attention matrix and obtain a vector of column-wise scores. 
Intuitively, the larger score indicates the stronger pattern that exists at the corresponding location. 
Thus, we select the maximum value of the column-wise score vector as the characteristic of knowledge aggregation patterns. 
Formally, 
\begin{equation}
\label{eq:4}
    \phi(\omega_{<t}) = \mathop{\prod}_{i=c}^{t-1}\sigma\omega_{i,c}, \quad\text{s.t. } c=\mathop{\arg\max}_{t-k\leq j\leq t-1}\mathop{\prod}_{i=j}^{t-1}\sigma\omega_{i,j}.
\end{equation}

Until now, we have an salient metric to detect the occurring of knowledge aggregation patterns within the local window. 
With the concern of calculation efficiency and the penalty should not bias the model to unreasonable output, we choose the top-$N_{can}$ in the logit of each beam to consist a candidate set $\mathcal{Y}$, where $|\mathcal{Y}|=N_{can}*N_{beam}$ and $N_{beam}$ is the number of beams. 
In this way, we limit the prediction within the candidate set and incorporate $\phi(w_{\leq t})$ with the model logits to predict the next token, \ie, 
\begin{equation}
    p(x_t|x_{<t}) = \text{Softmax}[\mathcal{H}(h_t) - \alpha\phi(w_{\leq t})]_{x_t}, \quad\text{s.t. } x_t\in \mathcal{Y},
\end{equation}
where $w_{\leq t}$ simplifies all of attention weights obtained by feeding forward the sequence $\{x_0,x_1,\ldots,x_t\}$.

\subsection{Retrospection-Allocation Strategy}
\label{sec:pas}

\begin{figure}[t]
\centering
\includegraphics[width=1.0\linewidth]{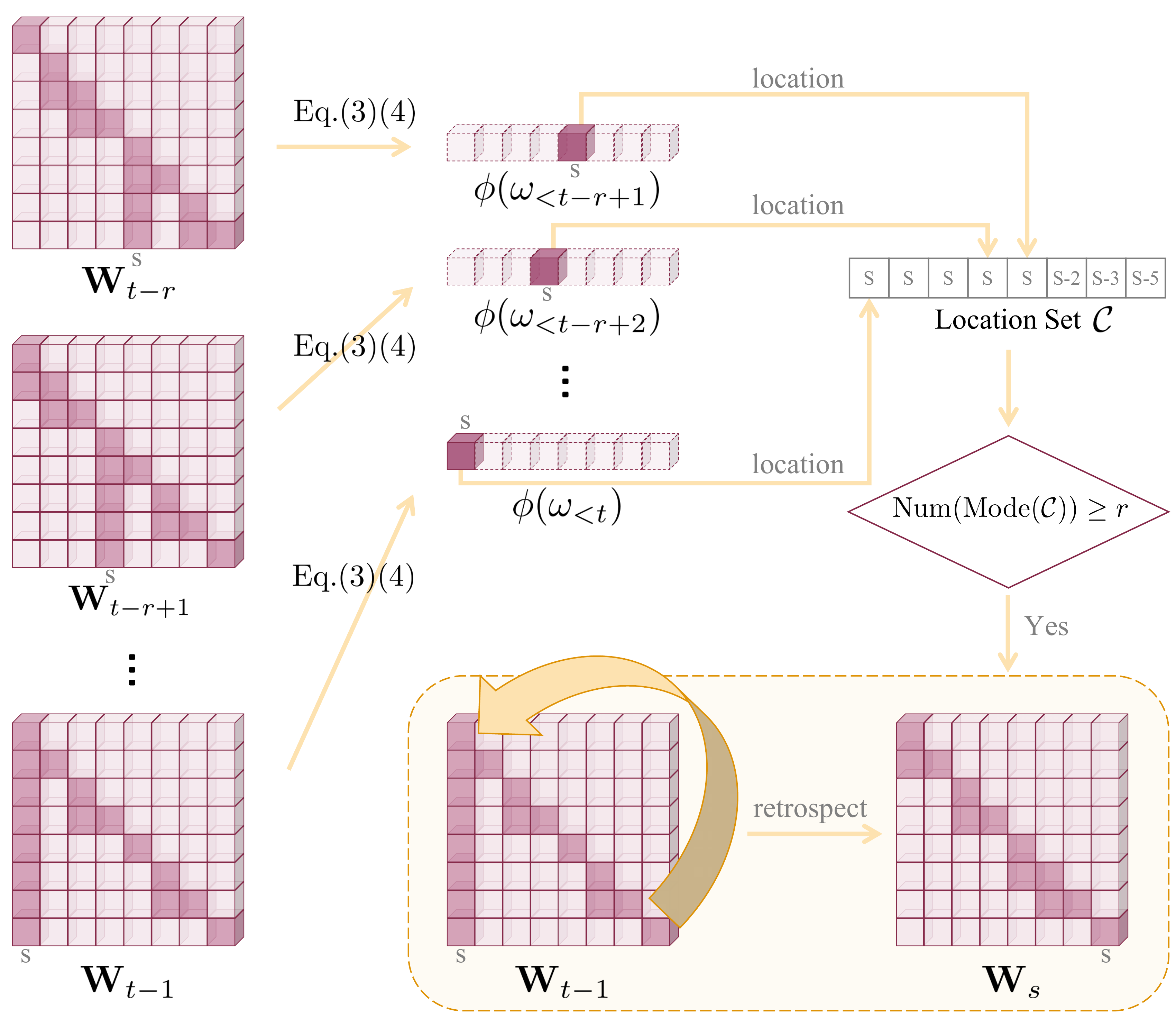}
\caption{The scheme of the proposed Retrospection strategy. We compute the maximum value coordinates of the past several token's column-wise scores and check if the overlap time is larger than $r$. If yes, we retrospect the decoding procedure and reselect the next token $x_{s+1}$.}
\label{fig:method2}
\end{figure}

With the over-trust logit penalty, we can successfully detect the occurrence of patterns after several subsequent tokens are generated. 
Normally, the penalty term is able to penalize the candidates which have knowledge aggregation patterns, and encourage other candidates to be predicted. While there still exists a few cases that  \textit{all of the candidates get penalized and the hallucination already occurred} 

This case motivates us to rethink the origin of such aggregation patterns: it is caused by the first few subsequent tokens over-trusting the summary token, and the penalty failed to correct them.
So an intuitive while aggressive idea is that the pattern will be greatly weakened if we could exclude the tokens that lead to hallucination and re-choose the proper first few tokens after the summary token. 

To this end, we propose the \textbf{R}etrospection-\textbf{A}llocation strategy. 
Specifically, when the decoding procedure encounters the knowledge aggregation pattern and the hallucination is inevitable, it rolls back to the summary token and selects other candidates for the next token prediction except for the candidates selected before. 
Empirically, the condition of decoding retrospection is designed as the location overlap of the maximum value in column-wise scores that corresponds to several consecutive tokens, where we manually set the threshold counts as $r$. Rather than the maximum value that varies between different models, location counting is a much more robust and general metric for the decision.

The whole retrospection process is illustrated in \Fref{fig:method2}. Based on \sref{sec:olp}, we can easily derive the location coordinate $c$ of the maximum score via \eref{eq:4}. 
Consequently, we can obtain the location coordinate set of several recently decoded tokens $x_{t-l},\ldots,x_{t-1}$, \ie,
\begin{equation}
    \mathcal{C} = \{c|c=\mathop{\arg\max}_{t-k\leq j\leq z}\mathop{\prod}_{i=j}^z\sigma\omega_{i,j}, z\in[t-l,t-1]\},
\end{equation}
where $l>r$ should be specified. We set $l=k$ by default.

Given a sequence $\{x_0,x_1,\ldots,x_{t-1}\}$ and its recent location coordinate set $\mathcal{C}$, we can easily check whether the coordinates are consistent. 
Formally, the overlap times can be calculated by
\begin{equation}
    N_{overlap} = \mathop{\sum}_{c\in\mathcal{C}} \mathbbm{1}_{c=s},\quad\text{s.t. }s=\text{Mode}(\mathcal{C}),
\end{equation}
where $\mathbbm{1}$ is an indicative function that returns 1 for the condition is true and returns 0 for the condition is false, $\text{Mode}$ is the function to get the mode of a set of values. 

If $N_{overlap}\geq r$, we consider to implement retrospection, regarding $s=\text{Mode}(\mathcal{C})$ as the location of the summary token. 
Suppose the sequence $\{x_0,x_1,\ldots,x_s,\ldots,x_{t-1}\}$ that has presented knowledge aggregation pattern at the summary token $x_s$, we intend to roll the decoding procedure back to the sequence $\{x_0,x_1,\ldots,x_s\}$ and select the new next token in the complementary set $\mathcal{Y}/\{x_{s+1}\}$. 
Since the subsequent rollback will be further forward than previous ones, we manually specify that the rollback location $s$ must be monotonically not decreasing. 
Additionally, we configure a maximum time $\beta$ for rollback and consider to roll back to $\{x_0,x_1,\ldots,x_{s-1}\}$ if $x_s$ has already reached the maximum rollback times.

\section{Experiment}

\subsection{Setup}

\noindent\textbf{Models.}
We select four of the most representative MLLM models for evaluation, including InstructBLIP \cite{instructblip}, MiniGPT-4 \cite{zhu2023minigpt}, LLaVA-1.5 \cite{liu2023improvedllava} and Shikra \cite{chen2023shikra}. 
These MLLM models can be roughly divided into two categories: 
Both InstructBLIP and MiniGPT-4 adopt Q-former \cite{li2023blip} to bridge the features between vision and text modality, using just 32 tokens to efficiently depict image representations. 
While LLaVA-1.5 and Shikra simply leverage linear projection layers to align the features of two modalities, with 256 or even 576 image tokens as MLLM input. 
All of these MLLM models apply a well-pretrained model as their vision encoder, such as CLIP \cite{radford2021clip} and EVA \cite{fang2023eva}, as well as a pretrained language model like LLaMA \cite{touvron2023llama} or Vicuna \cite{chiang2023vicuna}. 
Note that all of models used in our paper are 7B models.

\noindent\textbf{Baselines.} 
Since our work targets on the decoding approaches of MLLMs, we choose four decoding methods as the baseline methods, including three common strategies greedy decoding, Nucleus sampling, Beam search decoding and one method DoLa that is designed for mitigating LLMs' hallucination issues. 
Greedy decoding selects tokens step by step, greedily choosing the one with the highest probability in the language model logits.
Improved on greedy decoding, Beam search decoding \cite{graves2012sequence,sutskever2014sequence,boulanger2013audio} maintains a set of beams to enlarge the candidate range and select the best on in beams finally.
Different from the aforementioned two methods, nucleus sampling \cite{holtzman2019curious} concentrates concentrates on the predominant probability mass at each time step, maintaining a small subset of the vocabulary, typically ranging between one and a thousand candidates.
DoLa \cite{chuang2023dola}, designed for hallucination reduction in LLMs, contrasts the logits of the mature layer with those of pre-mature layers, using the increment as the final output logits. 
We adopt the default settings of all of these baseline methods, where we unify $N_{beam}=5$ for both Beam search and our OPERA, and set $p=0.9$ for nucleus sampling.
For DoLa, we use ``0,2,4,6,8,10,12,14'' as the indexes of candidate pre-mature layers and ``32'' as the index of the mature layer for DoLa.

\noindent\textbf{Implementation details.}
Basically, OPERA is established on Beam search where $N_{beam}=5$ by default. 
We empirically select $\sigma=50$ as the scaling factor in \eref{eq:4}, to ensure the attention values on knowledge aggregation patterns could be larger than 1 while the values on weaker attention areas could be smaller than 1. 
It aims to get the larger multiplication result on knowledge aggregation pattern. 
For the number $N_{can}$ of candidates, it is a configurable hyper-parameter like $N_{can}$ and we set $N_{can}=5$ by default. 
Too large $N_{can}$ will consume lots of time during decoding. 
Besides, we unify $\alpha=1$, $\beta=5$ and $r=15$ for all of MLLM models.

\begin{figure*}[h!]
\begin{minipage}{0.245\linewidth}
    \centering
    \includegraphics[width=1\linewidth]{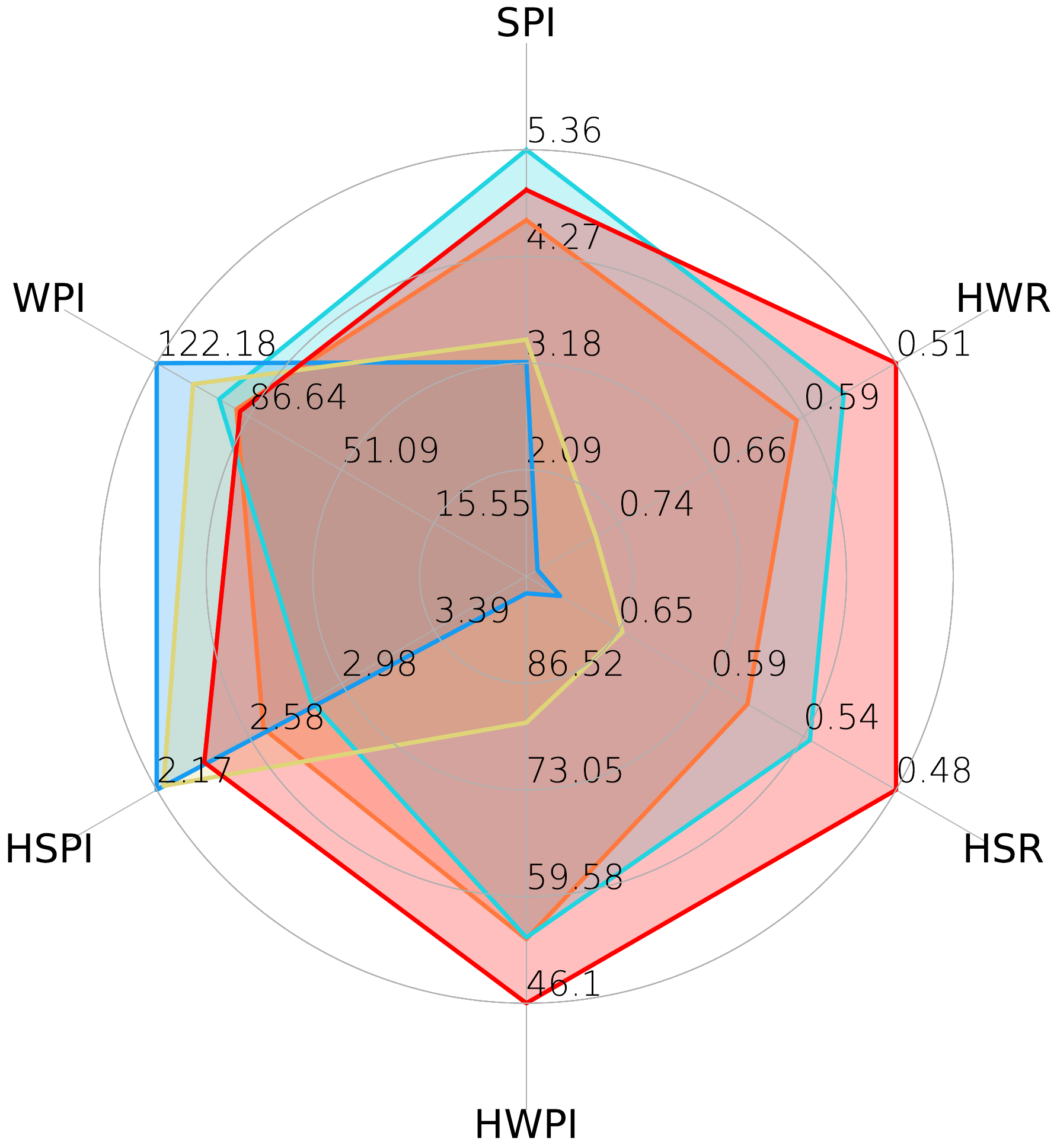}
    \\
    \scriptsize InstructBLIP
\end{minipage}
\hfill
\begin{minipage}{0.245\linewidth}
    \centering
    \includegraphics[width=1\linewidth]{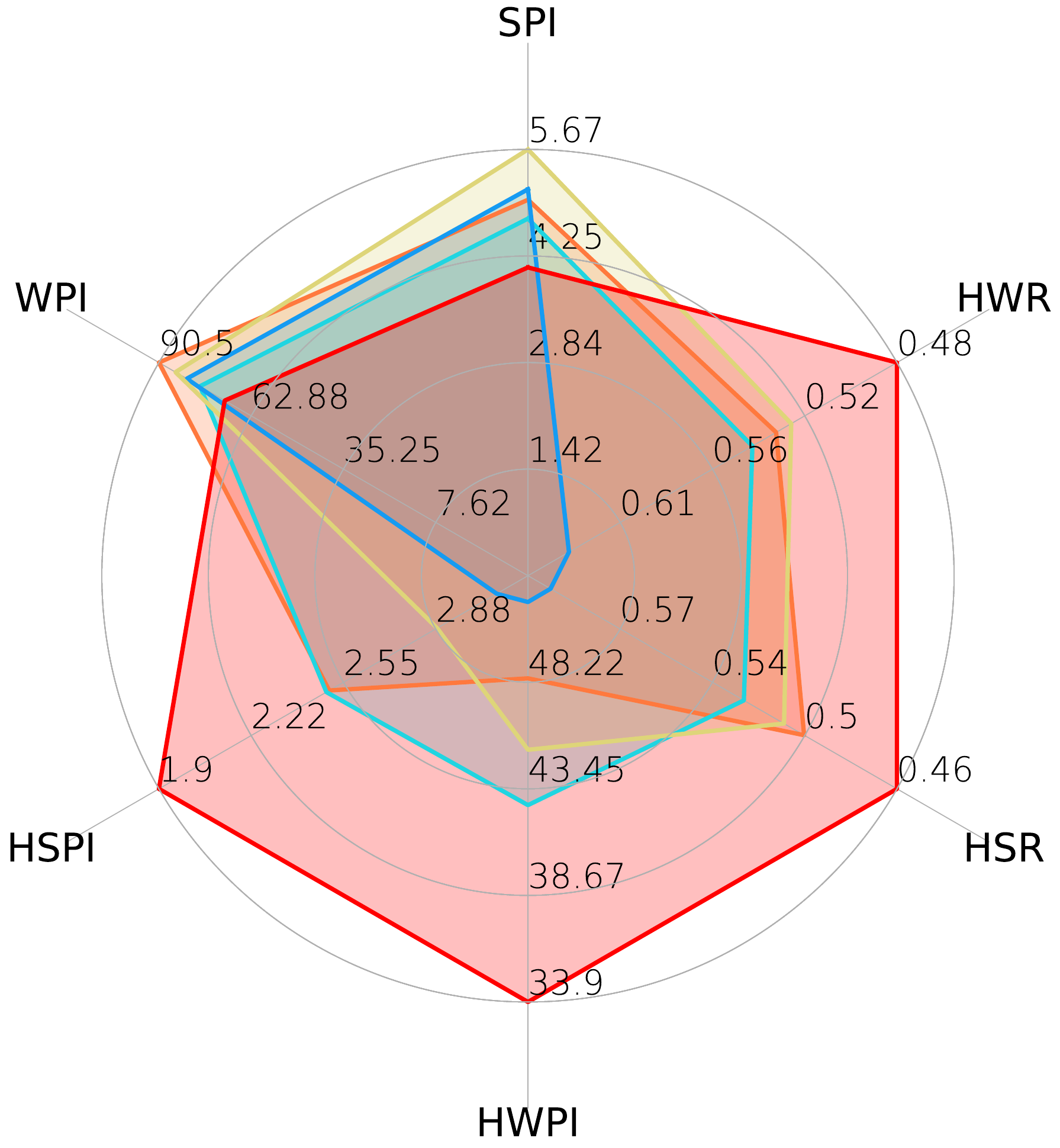}
    \\
    \scriptsize MiniGPT-4
\end{minipage}
\hfill
\begin{minipage}{0.245\linewidth}
    \centering
    \includegraphics[width=1\linewidth]{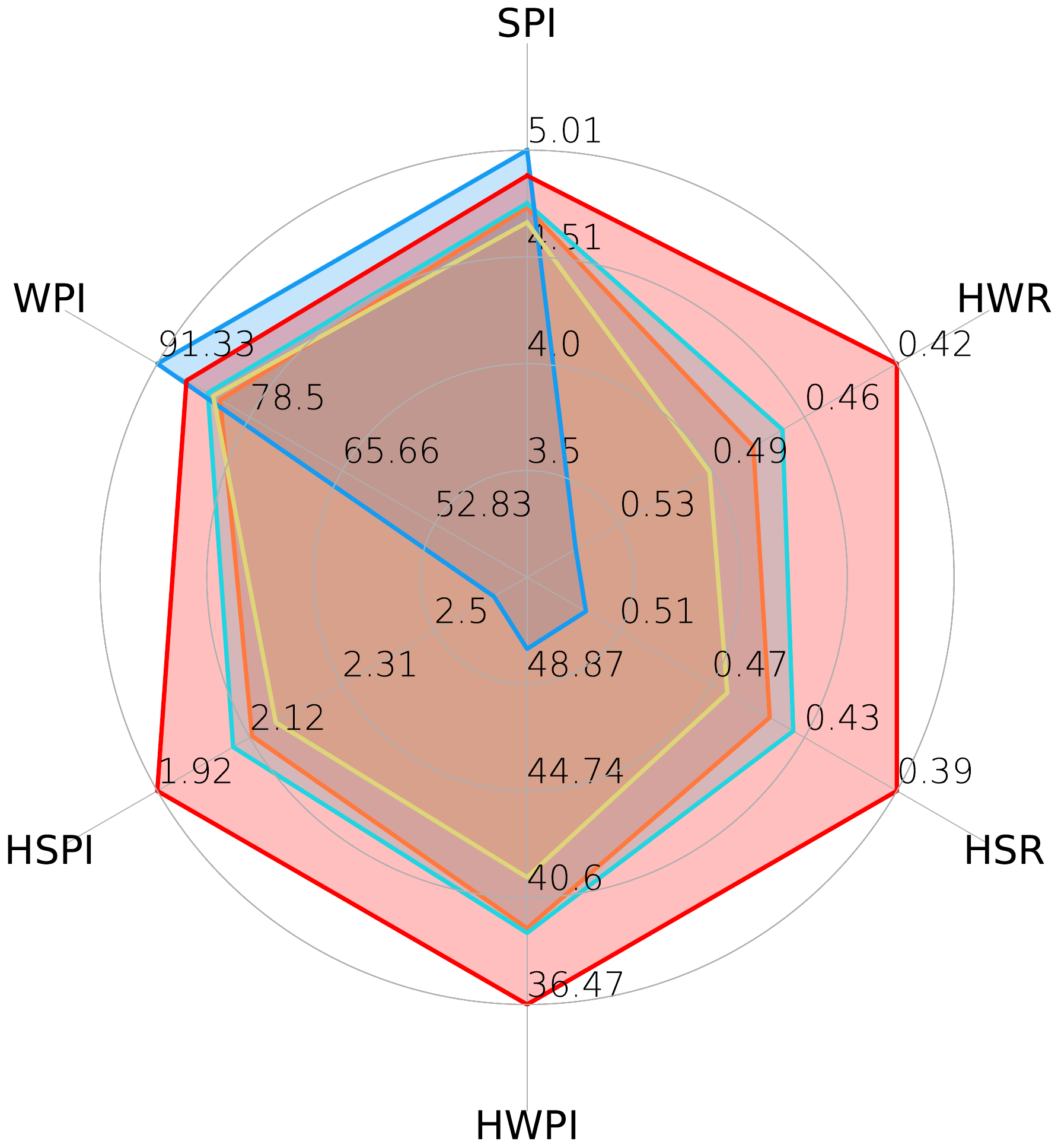}
    \\
    \scriptsize LLaVA-1.5
\end{minipage}
\hfill
\begin{minipage}{0.245\linewidth} 
    \centering
    \includegraphics[width=1\linewidth]{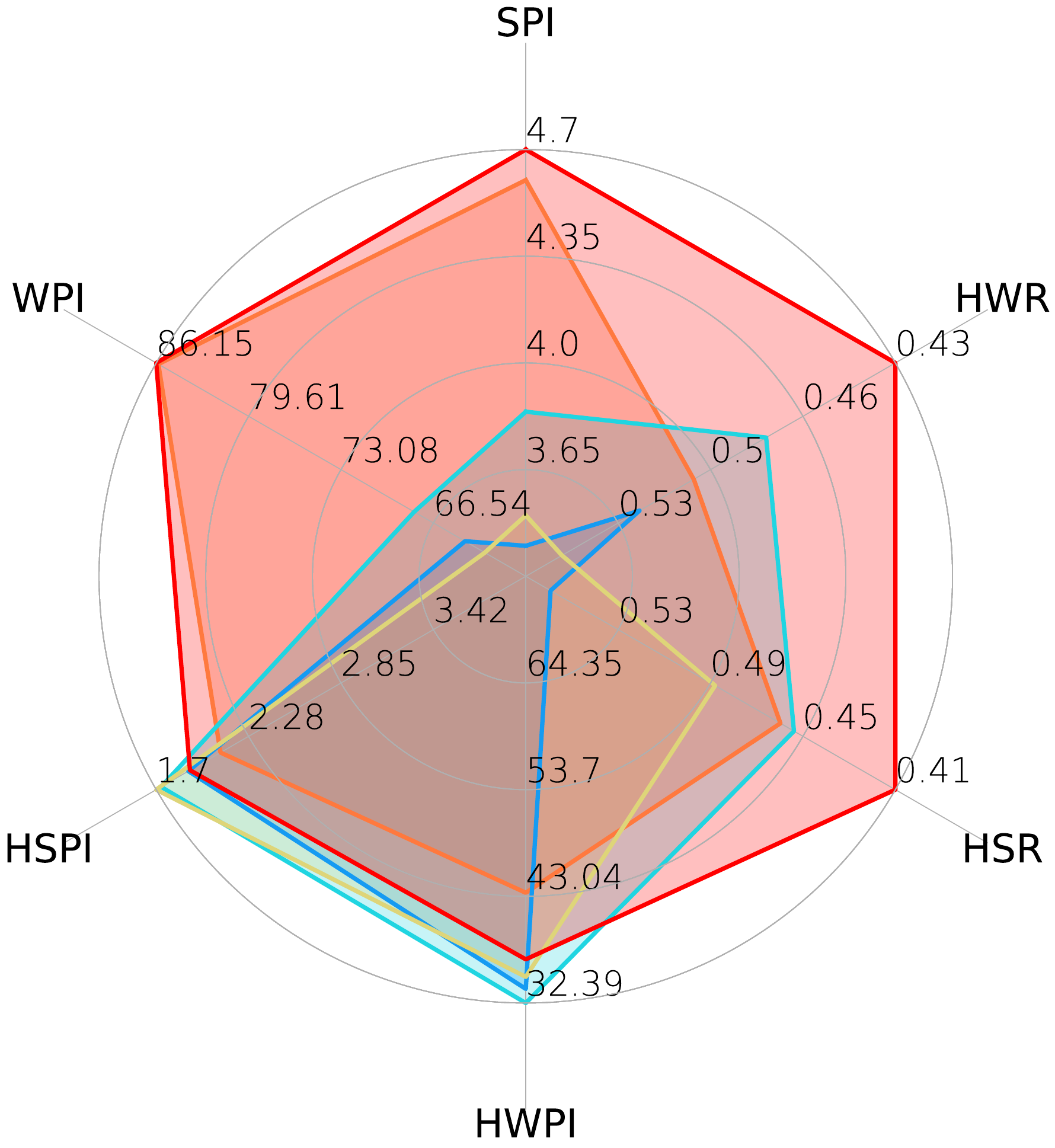}
    \\
    \scriptsize Shikra
\end{minipage}
\vfill
\centering
\includegraphics[width=1\linewidth]{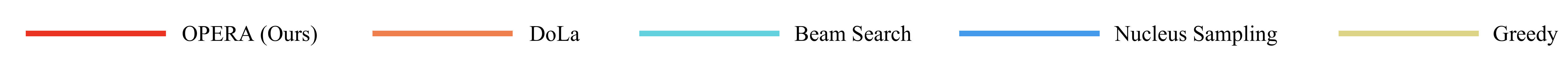}
\vspace{-2.5em}
\caption{GPT-4 assisted hallucination evaluation \cite{2023zhiyuan} results on VG-100K dataset. Six aspects of values are analyzed, including the number of sentences per image (SPI), the number of words per image (WPI), the number of hallucinated sentences per image (HSPI), the number of hallucinated words per image (HWPI), the ratio of hallucinated sentences (HSR), and the ratio of hallucinated words (HWR). Note that larger SPI and WPI, smaller HSPI, HWPI, HSR and HWR are better. Larger radar indicates better performance.}
\label{fig:radar}
\end{figure*}

\subsection{Quantitative Results}
In this section, we evaluate OPERA's performance of mitigating hallucinations on both long descriptions and simplified VQA answers. 

\noindent\textbf{CHAIR evaluation on hallucinations.}
\label{sec:chair_exp}
The Caption Hallucination Assessment with Image Relevance (CHAIR) \cite{rohrbach2018object} metric is a specifically crafted evaluation tool designed to assess object hallucination issues in image captioning task. 
More precisely, CHAIR quantifies the degree of object hallucination in a given image description by calculating the ratio of all objects mentioned in the description that are not present in the ground-truth label set.
It comprises two distinct assessment dimensions, including CHAIR$_S$ that calculates on sentence-level and CHAIR$_I$ that calculates on image-level. 
Denoted as $C_S$ and $C_I$, these two variants can be formulated as the average results of
\begin{equation}
    \scriptsize C_S = \frac{|\{\text{hallucinated objects}\}|}{|\{\text{all mentioned objects}\}|}, C_I = \frac{|\{\text{captions w/ hallucinated objects}\}|}{|\{\text{all captions}\}|},
\nonumber
\end{equation}
where the integration of CHAIR$_S$ and CHAIR$_I$ enables a thorough and detailed analysis of object hallucination issues in image captioning.

\begin{table}[t]
    \footnotesize
    \centering
    \setlength{\tabcolsep}{1mm}{
    \begin{tabular}{lp{6mm}<{\centering}p{6mm}<{\centering}p{6mm}<{\centering}p{6mm}<{\centering}p{6mm}<{\centering}p{6mm}<{\centering}p{6mm}<{\centering}p{6mm}<{\centering}}
        \toprule
        \multicolumn{1}{c}{\multirow{2}{*}{Method}}
        & \multicolumn{2}{c}{InstructBLIP} 
        & \multicolumn{2}{c}{MiniGPT-4} 
        & \multicolumn{2}{c}{LLaVA-1.5} 
        & \multicolumn{2}{c}{Shikra} 
        \\
        \cmidrule{2-9}
        & $C_S$ & $C_I$ & $C_S$ & $C_I$ & $C_S$ & $C_I$ & $C_S$ & $C_I$
        \\
        \midrule
        Greedy & 58.8 & 23.7 & 31.8 & 9.9 & 45.0 & 14.7 & 55.8 & 15.4 
        \\
        Nucleus & 54.6 & 24.8 & 32.6 & 10.7 & 48.8 & 14.2 & 55.6 & 15.4 
        \\
        Beam Search & 55.6 & 15.8 & 30.6 & 9.5 & 48.8 & 13.9 & 50.4 & 13.3 
        \\
        DoLa & 48.4 & 15.9 & 32.2 & 10.0 & 47.8 & 13.8 & 55.8 & 15.1 
        \\
        \textbf{OPERA (Ours)} & \textbf{46.4} & \textbf{14.2} & \textbf{26.2} & \textbf{9.5} & \textbf{44.6} & \textbf{12.8} & \textbf{36.2} & \textbf{12.1} 
        \\
        \bottomrule
    \end{tabular}
    }
    \vspace{-0.5em}
    \caption{CHAIR hallucination evaluation results on four MLLM models (\textit{max new tokens} is 512). Denote CHAIR$_S$ as $C_S$ and CHAIR$_I$ as $C_I$. Smaller values corresponds to less hallucinations.}
    \label{tab:chair_eval_512}
\end{table}

We conduct CHAIR evaluation on MSCOCO dataset \cite{lin2014microsoft}, which contains more than 300,000 images and 80 objects with annotations. 
Specifically, we randomly select 500 images in the validate set of COCO 2014 and query different MLLM models with the prompt ``\texttt{Please describe this image in detail.}'' to get their descriptions. 
Considering the length of sequences can greatly affect the values of CHAIR \cite{li2023pope}, we restrict two types of \textit{max new tokens} to generate descriptions for fair evaluation.

\begin{table}[t]
    \footnotesize
    \centering
    \setlength{\tabcolsep}{1mm}{
    \begin{tabular}{lp{6mm}<{\centering}p{6mm}<{\centering}p{6mm}<{\centering}p{6mm}<{\centering}p{6mm}<{\centering}p{6mm}<{\centering}p{6mm}<{\centering}p{6mm}<{\centering}}
        \toprule
        \multicolumn{1}{c}{\multirow{2}{*}{Method}}
        & \multicolumn{2}{c}{InstructBLIP} 
        & \multicolumn{2}{c}{MiniGPT-4} 
        & \multicolumn{2}{c}{LLaVA-1.5} 
        & \multicolumn{2}{c}{Shikra} 
        \\
        \cmidrule{2-3}
        \cmidrule{4-5}
        \cmidrule{6-7}
        \cmidrule{8-9}
        & $C_S$ & $C_I$ & $C_S$ & $C_I$ & $C_S$ & $C_I$ & $C_S$ & $C_I$
        \\
        \midrule
        Greedy & 30.0 & 14.5 & 24.2 & 8.2 & 20.6 & 6.2 & 22.0 & 7.0
        \\
        Nucleus & 30.4 & 15.7 & 23.6 & 8.3 & 26.2 & 8.5 & 22.6 & 7.6
        \\
        Beam Search & 21.4 & 7.2 & 23.6 & \textbf{7.8} & 18.8 & 5.9 & 20.2 & 6.4
        \\
        DoLa & 22.2 & 7.1 & 24.2 & 8.2 & 20.4 & 6.3 & 20.2 & 6.3
        \\
        \textbf{OPERA (Ours)} & \textbf{16.6} & \textbf{6.8} & \textbf{22.6} & {8.2} & \textbf{14.2} & \textbf{5.2} & \textbf{14.2} & \textbf{5.9}
        \\
        \bottomrule
    \end{tabular}
    }
    \vspace{-0.5em}
    \caption{CHAIR hallucination evaluation results on four MLLM models (\textit{max new tokens} is 64). Denote CHAIR$_S$ as $C_S$ and CHAIR$_I$ as $C_I$. Smaller values corresponds to less hallucinations.}
    \label{tab:chair_eval_64}
\end{table}

As shown in \Tref{tab:chair_eval_512} and \Tref{tab:chair_eval_64}, our OPERA obviously surpasses all of baselines decoding methods in both terms of $C_S$ and $C_I$. 
Especially on Shikra, our method achieves $\sim$35\% improvement on DoLa. 
The superior performances of OPERA are consistent between long description generation and short description generation.

\noindent\textbf{GPT-4 assisted evaluation.} 
CHAIR is a strong metric to evaluate the object-existence-level hallucination, while it fails to identify other kinds of hallucination, such as the attribute, location, and relation hallucination of objects.
HalluBench \cite{2023zhiyuan} is an advanced benchmark, which use the detailed object-level description in the VG dataset \cite{krishna2017visual} as ground-truth, and relay on the advanced GPT-4 to judge the hallucination in the description.
In practice, the detailed objects-level description are gathered as a disordered comprehensive description about the image, and the GPT-4 is carefully prompted to judge the hallucination in the MLLM generated descriptions, sentence by sentence.
Similar to \Sref{sec:chair_exp}, the MLLMs are prompted with the instruction ``\texttt{Please describe this image in detail.}'' and the \textit{max new tokens} is set to 512.
Details are shared in \Sref{sec:supp_4}.

From \Fref{fig:radar}, we observe that our OPERA generally achieves much less hallucinated sentences or words for describing each image, \eg, $\sim$30.4\% surpassing greedy decoding on the ratio of hallucinated sentences (HSR), and $\sim$15.4\% surpassing DoLa at the ratio of hallucinated words (HWR). 
It indicates that OPERA does help the model partially overcome the hallucination issue caused by its bias or over-trusting problems. 
We also notice that OPERA somehow slightly reduce the length of MLLM's output sequence, it is probably attributed by the reducing of those additional hallucinated contents.

\begin{table}[t]
\vspace{-0.5em}
    \footnotesize
    \centering
    \setlength{\tabcolsep}{1mm}{
    \begin{tabular}{lp{6mm}<{\centering}p{6mm}<{\centering}p{6mm}<{\centering}p{6mm}<{\centering}p{6mm}<{\centering}p{6mm}<{\centering}p{6mm}<{\centering}p{6mm}<{\centering}}
        \toprule
        \multicolumn{1}{c}{\multirow{2}{*}{Method}}
        & \multicolumn{2}{c}{InstructBLIP} 
        & \multicolumn{2}{c}{MiniGPT-4} 
        & \multicolumn{2}{c}{LLaVA-1.5} 
        & \multicolumn{2}{c}{Shikra} 
        \\
        \cmidrule{2-3}
        \cmidrule{4-5}
        \cmidrule{6-7}
        \cmidrule{8-9}
        & $C$ & $D$ & $C$ & $D$ & $C$ & $D$ & $C$ & $D$
        \\
        \midrule
        Beam Search & 5.52 & 5.26 & 5.29 & 5.06 & 5.53 & 5.15 & 5.25 & 5.08
        \\
        \textbf{OPERA (Ours)} & \textbf{6.26} & \textbf{5.27} & \textbf{6.87} & \textbf{5.08} & \textbf{6.32} & \textbf{5.16} & \textbf{6.29} & \textbf{5.26}
        \\
        \bottomrule
    \end{tabular}
    }
    \vspace{-0.5em}
    \caption{GPT-4V assisted hallucination evaluation results on MSCOCO. Two aspects are verified, \ie, correctness ($C$) and detailedness ($D$). Higher correctness indicates less hallucinations.}
    \label{tab:gpt_4v}
\end{table}

\begin{table}[t]
\vspace{-0.5em}
    \footnotesize
    \centering
    \setlength{\tabcolsep}{1mm}{
    \begin{tabular}{lp{14mm}<{\centering}p{14mm}<{\centering}p{14mm}<{\centering}p{14mm}<{\centering}}
        \toprule
        Method
        & InstructBLIP
        & MiniGPT-4
        & LLaVA-1.5
        & Shikra
        \\
        \midrule
        Greedy & 80.0 & 58.5 & 82.2 & 81.1
        \\
        Nucleus & 80.1 & 57.8 & 82.5 & 81.2
        \\
        Beam Search & 84.4 & 70.3 & 84.9 & 82.5
        \\
        DoLa & 83.4 & 72.8 & 83.2 & 82.1
        \\
        \textbf{OPERA (Ours)} & \textbf{84.8} & \textbf{73.3} & \textbf{85.4} & \textbf{82.7}
        \\
        \bottomrule
    \end{tabular}
    }
    \vspace{-0.5em}
    \caption{POPE hallucination evaluation results on four MLLM models. We report the average F1-score computed on \textit{random}, \textit{popular}, and \textit{adversarial} splits of POPE.}
    \label{tab:pope_f1}
\end{table}

\noindent\textbf{GPT-4V assisted evaluation.} 
We further resort to GPT-4Vision, a strong multi-modal assistant that can easily handle the input from vision, language, and voice modality. 
Typically, we randomly sample 500 images from MSCOCO's validate set and ask different MLLM models to describe these images. 
For fair comparison, we following \cite{yin2023woodpecker} and compare the answers obtained from two decoding methods at the same time, \ie, providing the image and both the answers to GPT-4V and prompting it to give a judgement from 0-10 respectively. 
The prompt emphasizes mitigating the impact of the sequential order fed to GPT-4V and, additionally, paying special attention to the objects mentioned in answers but not appear in the provided image. 
This includes instances where the objects are represented in an incorrect form, such as wrong colors, positions, or relationships.
Details are shared in \Sref{sec:supp_5}.

As showcased in \Tref{tab:gpt_4v}, our OPERA achieves up to 27.5\% improvements compared with Beam search decoding, while keeping the detailedness of answers. 
Since GPT-4V's abilities of perception and reasoning are very closed to human beings, the GPT-4V evaluation results somehow reflect the strong performance of reducing hallucinations from the perspective of human's feeling.

\noindent\textbf{POPE evaluation on hallucinations.} 
The Polling-based Object Probing Evaluation (POPE) \cite{li2023pope} is a recently introduced method designed to assess hallucination issues in MLLMs. 
Similar to CHAIR, POPE focuses on evaluating object hallucination, utilizing an essay question format to prompt the model like ``\texttt{Is There a <object> in the image?}'', to determine whether the model can configure out the given image corresponds to a specific object.
The complete POPE test comprises three splits:
In the``random'' split, the evaluation randomly selects objects from the whole dataset.
In the ``popular'' split, the evaluation assesses the presence of objects that most frequently appear in the dataset.
In the ``adversarial'' split, it evaluates the MLLM's ability to identify objects highly relevant to those present in the image.

We verify POPE on four MLLM models and report the average F1 scores in \Tref{tab:pope_f1}. 
Compared with baseline methods, we can observe our OPERA also attains the highest performance among these decoding strategies, albeit with marginal gains. 
It is essential to clarify that our approach excels specifically in alleviating hallucinations within \textbf{lengthy sequences}. 
In the context of POPE answers, where responses typically start with \texttt{Yes} or \texttt{No} and conclude as quite brief sequences like ``\texttt{Yes, there is a <object> in the image.}'', the knowledge aggregation patterns, a crucial hypothesis of our method, may not manifest as prominently.

\begin{table}[t]
\vspace{-0.5em}
    \footnotesize
    \centering
    \setlength{\tabcolsep}{3mm}{
    \begin{tabular}{l|p{7mm}<{\centering}p{6mm}<{\centering}p{7mm}<{\centering}p{7mm}<{\centering}p{7mm}<{\centering}}
        \toprule
        & PPL$_1$$\downarrow$ & PPL$_2$$\downarrow$& Grammar$\uparrow$ & Fluency$\uparrow$ & Natural$\uparrow$
        \\
        \midrule
        Greedy & 12.72 & 10.27 & \textbf{9.58} & \textbf{9.01} & 8.52
        \\
        Nucleus & 17.17 & 13.78 & 8.51 & 8.53 & 7.95
        \\
        Beam Search & \textbf{11.11} & \textbf{8.89} & \underline{9.54} & \underline{8.95} & \textbf{8.55}
        \\
        DoLa & 12.89 & 10.40 & 9.31 & 8.89 & 8.46
        \\
        OPERA & \underline{11.67} & \underline{9.31} & \underline{9.54} & 8.93 & \underline{8.53}
        \\
        \bottomrule
    \end{tabular}
    }
    \vspace{-0.5em}
    \caption{The evaluation results for the quality of generated text. We calculate PPL$_1$ and PPL$_2$ with \textit{gpt2} and \textit{gpt2-medium} in the \textit{huggingface} model zoo respectively. The ratings of grammer, fluency, and naturalness is given by GPT-4.}
    \label{tab:ppl}
\end{table}

\begin{table}[t]
\vspace{-0.5em}
    \footnotesize
    \centering
    \setlength{\tabcolsep}{3mm}{
    \begin{tabular}{l|p{7mm}<{\centering}p{7mm}<{\centering}p{7mm}<{\centering}p{7mm}<{\centering}p{7mm}<{\centering}}
        \toprule
        & Greedy & Nucleus & Beam & DoLa & \textbf{OPERA}
        \\
        \midrule
        MMBench & 64.3 & 64.0 & 64.4 & 63.8 & \textbf{64.4}
        \\
        MME & 1510.7 & 1471.9 & 1504.3 & 1480.1 & \textbf{1515.4}
        \\
        \bottomrule
    \end{tabular}
    }
    \vspace{-0.5em}
    \caption{OPERA generally improves the MLLM's performance on popular MLLM benchmark.}
    \label{tab:mllm_bench}
\end{table}

\begin{table*}[t]
    \footnotesize
    \centering
    \setlength{\tabcolsep}{1mm}{
    \begin{tabular}{p{15mm}<{\centering}p{9mm}<{\centering}p{9mm}<{\centering}p{9mm}<{\centering}p{9mm}<{\centering}p{9mm}<{\centering}p{9mm}<{\centering}p{9mm}<{\centering}p{9mm}<{\centering}p{9mm}<{\centering}p{9mm}<{\centering}p{9mm}<{\centering}p{9mm}<{\centering}}
        \toprule
        \multicolumn{1}{c}{\multirow{2}{*}{Setting}}
        & \multicolumn{1}{c}{\multirow{2}{*}{$N_{can}$}}
        & \multicolumn{1}{c}{\multirow{2}{*}{$\sigma$}}
        & \multicolumn{1}{c}{\multirow{2}{*}{$\alpha$}}
        & \multicolumn{1}{c}{\multirow{2}{*}{$r$}}
        & \multicolumn{2}{c}{InstructBLIP} 
        & \multicolumn{2}{c}{MiniGPT-4} 
        & \multicolumn{2}{c}{LLaVA-1.5} 
        & \multicolumn{2}{c}{Shikra} 
        \\
        \cmidrule{6-7}
        \cmidrule{8-9}
        \cmidrule{10-11}
        \cmidrule{12-13}
        & & & & & $C_S$ & $C_I$ & $C_S$ & $C_I$ & $C_S$ & $C_I$ & $C_S$ & $C_I$
        \\
        \midrule
        Beam Search & - & - & - & - & 55.6 & 15.8 & 30.6 & 9.5 & 48.8 & 13.9 & 50.4 & 13.3
        \\
        \midrule
        $\mathbf{A1}$ & 2 & 50 & 1 & 15 & \textbf{43.8} & \textbf{13.1} & 29.8 & 10.8 & \textbf{41.2} & \textbf{12.0} & 43.0 & 12.8
        \\
        $\mathbf{A2}$ & 3 & 50 & 1 & 15 & 46.4 & 13.2 & 30.0 & 10.0 & 43.8 & 12.8 & 39.4 & 12.7
        \\
        $\mathbf{A3}$ & 5 & 50 & 1 & 15 & 46.4 & 14.2 & \textbf{26.2} & \textbf{9.5} & 44.6 & 12.8 & 36.2 & 12.1 
        \\
        $\mathbf{A4}$ & 8 & 50 & 1 & 15 & 49.6 & 14.6 & 29.0 & 10.1 & 49.0 & 13.4 & \textbf{33.3} & \textbf{11.5}
        \\
        $\mathbf{A5}$ & 10 & 50 & 1 & 15 & 51.4 & 15.0 & 30.4 & 10.0 & 48.0 & 13.2 & 34.4 & 11.6
        \\
        \midrule
        $\mathbf{B1}$ & 5 & 40 & 1 & 15 & 47.6 & 14.3 & 27.8 & 10.2 & 46.9 & 13.3 & 45.4 & 12.8
        \\
        $\mathbf{B2}$ & 5 & 45 & 1 & 15 & 47.2 & 14.5 & 26.8 & 9.8 & 47.8 & 13.3 & 41.2 & 12.3
        \\
        $\mathbf{B3}$ & 5 & 50 & 1 & 15 & 46.4 & 14.2 & 26.2 & 9.5 & \textbf{44.6} & \textbf{12.8} & 36.2 & 12.1 
        \\
        $\mathbf{B4}$ & 5 & 55 & 1 & 15 & 44.2 & \textbf{13.9} & \textbf{25.6} & \textbf{9.2} & 47.5 & 13.3 & 35.4 & 11.7
        \\
        $\mathbf{B5}$ & 5 & 60 & 1 & 15 & \textbf{44.0} & 14.3 & 26.6 & 10.9 & 44.5 & 13.0 & \textbf{33.8} & \textbf{11.7}
        \\
        \midrule
        $\mathbf{C1}$ & 5 & 50 & 0.1 & 15 & 47.6 & 14.4 & 26.6 & 9.7 & 46.4 & 12.8 & 40.2 & 12.4
        \\
        $\mathbf{C2}$ & 5 & 50 & 0.5 & 15 & 46.2 & 14.3 & 27.6 & 9.7 & 46.4 & 13.3 & 35.6 & \textbf{11.5}
        \\
        $\mathbf{C3}$ & 5 & 50 & 1 & 15 & 46.4 & 14.2 & \textbf{26.2} & \textbf{9.5} & \textbf{44.6} & \textbf{12.8} & 36.2 & 12.1 
        \\
        $\mathbf{C4}$ & 5 & 50 & 5 & 15 & 46.0 & \textbf{13.8} & 27.2 & 9.9 & 47.6 & 13.5 & 39.2 & 13.2
        \\
        $\mathbf{C5}$ & 5 & 50 & 10 & 15 & \textbf{45.4} & 14.0 & 26.4 & 9.5 & 46.4 & 13.2 & \textbf{35.4} & 12.6
        \\
        \midrule
        $\mathbf{D1}$ & 5 & 50 & 1 & 5 & 52.0 & 14.8 & \textbf{24.9} & 9.8 & 45.0 & 13.0 & 40.2 & 12.7
        \\
        $\mathbf{D2}$ & 5 & 50 & 1 & 10 & 50.4 & 14.8 & 26.4 & 10.1 & 45.3 & 12.9 & \textbf{36.0} & \textbf{11.5}
        \\
        $\mathbf{D3}$ & 5 & 50 & 1 & 15 & 46.4 & 14.2 & 26.2 & \textbf{9.5} & \textbf{44.6} & \textbf{12.8} & 36.2 & 12.1 
        \\
        $\mathbf{D4}$ & 5 & 50 & 1 & 20 & 42.6 & 13.4 & 27.1 & 9.7 & 45.6 & 13.0 & 37.0 & 12.1
        \\
        $\mathbf{D5}$ & 5 & 50 & 1 & 25 & \textbf{41.8} & \textbf{13.1} & 27.6 & 9.8 & 45.0 & 12.9 & 40.0 & 13.3
        \\
        \bottomrule
    \end{tabular}
    }
    \caption{Ablation studies on the hyper-parameters used in our OPERA, including the number of candidates $N_{can}$, the scale factor $\sigma$, the penalty weight $\alpha$ and the rollback threshold $r$. Denote CHAIR$_S$ as $C_S$ and CHAIR$_I$ as $C_I$. Lower values mean less hallucinations.}
    \label{tab:abl_hp}
\end{table*}

\noindent\textbf{Text quality evaluation.} 
To assess the overall quality of generated text comprehensively, we adopt PPL (Perplexity, a classical metric in NLP without using reference text), and resort to GPT-4 to assess the grammar, fluency, and naturalness of generated text. 
We randomly select 1,000 images in MSCOCO and verify on LLaVA-1.5 7B model. 
The average results are listed above, where PPL$_1$ and PPL$_2$ are calculated by pretrained \textit{gpt2} and \textit{gpt2-medium} respectively.

From the results in \Tref{tab:ppl}, we discover that OPERA can generally keep the quality of generated text from various aspects.
Besides, we test OPERA on two popular MLLM benchmark, \ie, MME \cite{fu2023mme} and MMBench \cite{liu2023mmbench}, using LLaVA-1.5 7B model. \Tref{tab:mllm_bench} shows that OPERA can maintain and even improve MLLM's performance on both MLLM benchmarks.

\begin{table}[t]
    \footnotesize
    \centering
    \setlength{\tabcolsep}{1mm}{
    \begin{tabular}{p{3mm}<{\centering}p{4mm}<{\centering}p{4mm}<{\centering}p{6mm}<{\centering}p{6mm}<{\centering}p{6mm}<{\centering}p{6mm}<{\centering}p{6mm}<{\centering}p{6mm}<{\centering}p{6mm}<{\centering}p{6mm}<{\centering}}
        \toprule
        \multicolumn{1}{c}{\multirow{2}{*}{Setup}}
        & \multicolumn{1}{c}{\multirow{2}{*}{P}}
        & \multicolumn{1}{c}{\multirow{2}{*}{R}}
        & \multicolumn{2}{c}{InstructBLIP} 
        & \multicolumn{2}{c}{MiniGPT-4} 
        & \multicolumn{2}{c}{LLaVA-1.5} 
        & \multicolumn{2}{c}{Shikra} 
        \\
        \cmidrule{4-5}
        \cmidrule{6-7}
        \cmidrule{8-9}
        \cmidrule{10-11}
        & & & $C_S$ & $C_I$ & $C_S$ & $C_I$ & $C_S$ & $C_I$ & $C_S$ & $C_I$
        \\
        \midrule
        $\mathbf{A}$ & \XSolidBrush & \XSolidBrush & 55.6 & 15.8 & 30.6 & 9.5 & 48.8 & 13.9 & 50.4 & 13.3 
        \\
        $\mathbf{B}$ & \XSolidBrush & \Checkmark & 50.0 & 14.6 & 27.3 & 10.1 & 46.4 & 12.9 & 46.8 & 13.0
        \\
        $\mathbf{C}$ & \Checkmark & \XSolidBrush & 48.2 & \textbf{13.8} & 27.4 & 10.0 & 45.2 & 13.0 & 41.8 & 13.9
        \\
        $\mathbf{D}$ & \Checkmark & \Checkmark & \textbf{46.4} & 14.2 & \textbf{26.2} & \textbf{9.5} & \textbf{44.6} & \textbf{12.8} & \textbf{36.2} & \textbf{12.1} 
        \\
        \bottomrule
    \end{tabular}
    }
    \vspace{-0.5em}
    \caption{Ablation results on two components. ``P'' denotes the over-trust penalty, ``R'' denotes retrospection-reallocation strategy.}
    \label{tab:abl_components}
\end{table}

\begin{table*}[t]
\footnotesize
\centering
\begin{tabular}{p{160mm}}
    \toprule
    GPT-4 Prompt
    \\
    \midrule
    Please help me judge if the comment of this image is hallucination or correct. 
    \\
    I will give you a list of region description of a image. The format is [x1, y1, x2, y2]: region description, where [x1, y1, x2, y2] is the bounding box of the region. Highly overlapping bounding boxes may refer to the same object. This is the ground truth information of the image. Your judgement should base on this information. However, this information only describe the objects in the region of image, so it cannot describe the subjective part of the image, e.g., atmosphere, style, emotion. In that case, you can return ``Cannot judge''. 
    \\
    Also, I will give you a list of comments of the image for you to judge if it is hallucination. Please give a judgement one by one along with the reason.
    \\
    You should pay extra attention to the hallucination, which refers to the part of comments that are inconsistent with the descriptions, specially claiming the existence of something not present in the descriptions.
    \\ \\
    If a comment is hallucination, please help me rewrite it. When rewrite the comment, sound like you are looking at the image directly.
    \\
    Each rewritten comments should compose a description about the image which is correct, detailed, smooth and has strong readability. 
    \\
    If not hallucination (correct or cannot judge), keep the original comment.
    \\ \\
    Your output should be:\\
    Judgement:\\
    1. hallucination or correct or cannot judge: $<$reason$>$\\
    2. ...\\
    Revised Sentences:\\
    1. ...\\
    2. ...\\ \\
    Here are the region descriptions of the image:\\
    $\{\}$
    \\
    Here is the comment for you to judge if it is hallucination and revise: \\
    $\{\}$
    \\
    \bottomrule
\end{tabular}
\caption{The prompt used for GPT-4 evaluation.}
\label{tab:gpt4_prompt}
\end{table*}

\subsection{Ablation Study on Hyper-parameters}
\label{sec:supp_2}

In this section, we give detailed ablation studies for hyper-parameters, including two key components, the number of candidates $N_{can}$, the scale factor $\sigma$, the penality weight $\alpha$, and the threshold $r$ of retrospection. 
Despite the best parameter of different MLLMs are a little bit different, OPERA is generally robust on the varying settings of hyper-parameters and outperforms the baselines. 
In our paper, we simply adopt a default setting with $N_{can}=5$, $\sigma=50$, $\alpha=1$, and $r=15$ for all MLLMs.

\noindent\textbf{Key components.} 
Here we ablate the two components proposed in OPERA, \ie, the over-trust penalty and the retrospection-reallocation strategy. 
As the results shown in \Tref{tab:abl_components}, when we discard both components, our method degrade to standard Beam search and presents worst perfoemance. 
Equipped either of the two components can help MLLM models hallucinate less, where the over-trust penalty contributes relatively more to the final performance. 
It is promising, since not all of generated sequences need to retrospect during decoding, unless encountering the knowledge aggregation patterns. 

\noindent\textbf{Number of candidates $N_{can}$.}
To prevent the model give unreasonable output, we restrict the prediction of each beam within the top-$N_{can}$ highest vocabularies in the logit. 
Note that $N_{can}$ is a configurable parameter like $N_{beam}$ in Beam Search \cite{graves2012sequence,sutskever2014sequence,boulanger2013audio}. 
An appropriate setup of $N_{can}$ can greatly improve the performance of OPERA. 
Too small $N_{can}$ may decrease the effect of retrospection-reallocation, while too large $N_{can}$ probably engages some unreasonable vocabularies that are irrelevant with the whole sequence. 
The results are listed in \Tref{tab:abl_hp}. 
InstructBLIP \cite{instructblip} and LLaVA-1.5 \cite{liu2023improvedllava} may prefer smaller $N_{can}$, while MiniGPT-4 \cite{zhu2023minigpt} prefers $N_{can}=5$ and Shikra \cite{chen2023shikra} prefers larger $N_{can}$.

\noindent\textbf{Scale Factor $\sigma$.} 
Before depicting the knowledge aggregation pattern through column-wise multiplication in attention maps, we set a scale factor $\sigma$ to scale up attention values which are usually too small. 
As the results presented in \Tref{tab:abl_hp}, different MLLM models prefer different scale factors, probably because the varying sequence lengths (\eg, LLaVA-1.5-7B has 576 image tokens while MiniGPT-4-7B has only 32 image tokens) result in different magnitudes of self-attention weight values (Note that the sum of self-attention weights should be 1). 
In other words, $\sigma$ is a configurable parameter for users to pursue the best performance of their own MLLM model in the rough range of 40 to 60. 
For simplicity, we set $\sigma$ as 50, a balanced choice that performs not bad on different MLLMs.

\noindent\textbf{Penalty weight $\alpha$.} 
We further ablate the weight of the introduced penalty term that is incorporated with the model logit. 
From the results in \Tref{tab:abl_hp}, we can observe that OPERA's performance is relatively robust when $\alpha$ varies. 
Different MLLMs may prefer different $\alpha$, but the numerical fluctuations are generally slight. 
For simplicity, we unify $\alpha$ as 1 for different MLLMs.

\noindent\textbf{Rollback threshold $r$.} 
We consider the location overlap of the maximum column-wise scores of several consecutive tokens as the condition of retrospection, where we set a threshold $r$ for the count of overlap. 
If the count of overlap reaches the threshold $r$, the rollback will be triggered. 
Consequently, the choice of $r$ seems crucial and a ablation study is necessary. 
The abaltion results are shown in \Tref{tab:abl_hp}. 
We can observe that InstructBLIP shows less hallucinations when $r=25$ while the other three MLLMs show have the better perofrmance when $r=15$. 
Therefore, we assign $r$ as 15 by default.

\begin{table*}[t]
\vspace{2em}
\footnotesize
\centering
\begin{tabular}{p{160mm}}
    \toprule
    GPT-4V(ision) Prompt
    \\
    \midrule
    You are required to score the performance of two AI assistants in describing a given image. You should pay extra attention to the hallucination, which refers to the part of descriptions that are inconsistent with the image content, such as claiming the existence of something not present in the image or describing incorrectly in terms of the counts, positions, or colors of objects in the image. Please rate the responses of the assistants on a scale of 1 to 10, where a higher score indicates better performance, according to the following criteria:\\
    1: Accuracy: whether the response is accurate with respect to the image content. Responses with fewer hallucinations should be given higher scores.\\
    2: Detailedness: whether the response is rich in necessary details. Note that hallucinated descriptions should not countas necessary details.\\
    Please output the scores for each criterion, containing only two values indicating the scores for Assistant 1 and 2, respectively. The two scores are separated by a space. Following the scores, please provide an explanation of your evaluation, avoiding any potential bias and ensuring that the order in which the responses were presented does not affect your judgment.\\ \\
    \text{[Assistant 1]}\\
    $\{\}$\\
    \text{[End of Assistant 1]}\\ \\
    \text{[Assistant 2]}\\
    $\{\}$\\
    \text{[End of Assistant 2]}\\ \\
    Output format:\\
    Accuracy: $<$Scores of the two answers$>$\\
    Reason:\\ \\
    Detailedness: $<$Scores of the two answers$>$\\
    Reason: \\ \\
    \bottomrule
\end{tabular}
\caption{The prompt used for GPT-4V(ision) evaluation.}
\label{tab:gpt4v_prompt}
\end{table*}


\subsection{Details of GPT-4 Evaluation}
\label{sec:supp_4}

We generally follow the GPT-4 evaluation proposed in HalluBench \cite{2023zhiyuan} and implement it on VG dataset. 
Each image in VG \cite{krishna2017visual} dataset has the detailed ground-truth descriptions about all of the appearing objects. 
Since GPT-4 is not able to deal with image data, we integrate all of ground-truth descriptions into the input prompt to help GPT-4 comprehend the image content. 
Then, given the MLLM's generated description on the image with ``\texttt{Please describe this image in detail.}'', GPT-4 are required to judge whether each sentences of MLLM's description has hallucinated contents. 
This evaluation is quite strict, where GPT-4 judges any MLLM's descriptions as hallucinations if they are deviated from the ground-truth descriptions in terms of quantity, color, location, activity, or direction.

\noindent\textbf{Metrics.} 
There are six metrics considered, which include: 
\begin{itemize}
    \item \textit{The number of sentences per image (SPI).} It reflects the detailedness of MLLM's description at the sentence level. 
    \item \textit{The number of words per image (WPI).} It reflects the detailedness of MLLM's description at the word level. 
    \item \textit{The number of hallucinated sentences per image (HSPI).} It reveals the hallucination degree of MLLM's description at the sentence level. Any sentences that contain hallucinated contents are taken into calculation. 
    \item \textit{The number of hallucinated words per image (HWPI).} It reveals the hallucination degree of MLLM's description at the word level. Any words related with hallucinated contents are taken into calculation. 
    \item \textit{The ratio of hallucinated sentences (HSR).} The average ratio of hallucinated sentences in all sentences of MLLM's descriptions on different images.
    \item \textit{The ratio of hallucinated words (HWR).} The average ratio of hallucinated words in all words of MLLM's descriptions on different images.
\end{itemize}

\noindent\textbf{Prompt.} 
As shown in \Tref{tab:gpt4_prompt}, our adopted GPT-4 prompt is generally based on HalluBench \cite{2023zhiyuan}.

\begin{figure*}[t]
\centering
\includegraphics[width=1.0\linewidth]{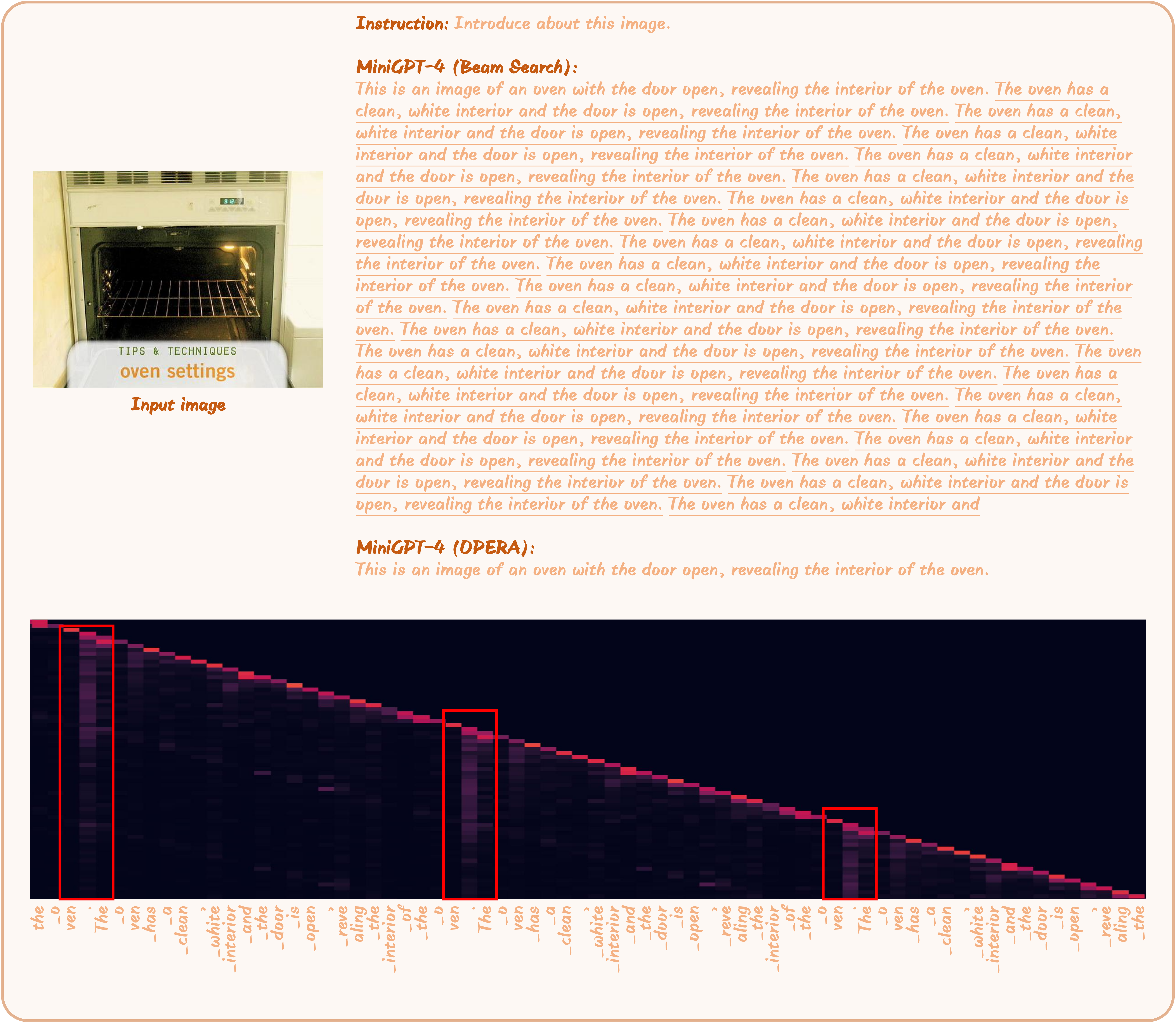}
\caption{OPERA's potentials for eliminating repetition. }
\label{fig:repetition_case}
\end{figure*}

\subsection{Details of GPT-4v Evaluation}
\label{sec:supp_5}

Following \cite{yin2023woodpecker}, we conduct the dual evaluation on GPT-4V(ision) for Beam search and our proposed OPERA. 
Given a trained MLLM model and a image, we respectively use Beam search decoding and OPERA decoding to obtain two descriptions with the prompt ``\texttt{Please describe this image in detail.}''. 
Then, we adopt the prompt shown in \Tref{tab:gpt4v_prompt} to ask GPT-4V to rate the two description based on the image on a scale of 0 to 10, where the rating involves two aspects, \ie, Accuracy and Detailedness. 
The accuracy reflects the consistency between the description and the given image.
If GPT-4V thinks any content in this description is inconsistent with the given image, namely higher hallucinations, it will get lower score. 
The detailedness reflects the degree of expressive ability, \ie, how comprehensive does the description characterize the image.

The prompt adopted for GPT-4V is listed in \Tref{tab:gpt4v_prompt}. 
It requires GPT-4V to ignore the bias incurred by the sequntial order and pay extra attention to the objects mentioned by MLLM's descriptions but not appear in the image, including incorrect colors, positions, or relationships. 
GPT-4V comprehensively analyzes MLLM's description, using its strong abilities that are closed to human.

\begin{figure}[t]
\centering
\includegraphics[width=1.0\linewidth]{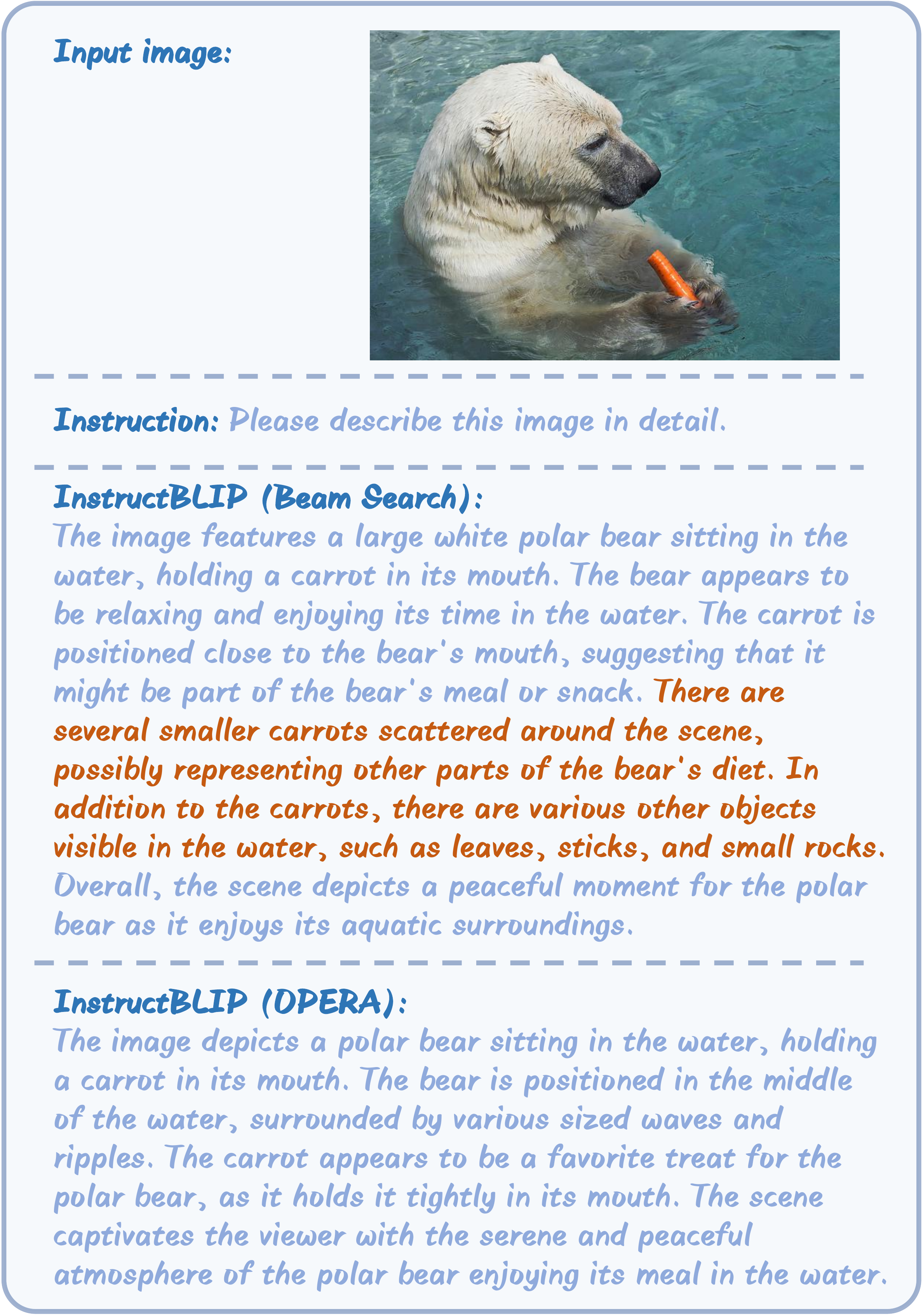}
\vspace{-1em}
\caption{OPERA's performance on reducing hallucinations. }
\label{fig:cases}
\vspace{-0.5em}
\end{figure}

\begin{figure*}[t]
\centering
\includegraphics[width=1.0\linewidth]{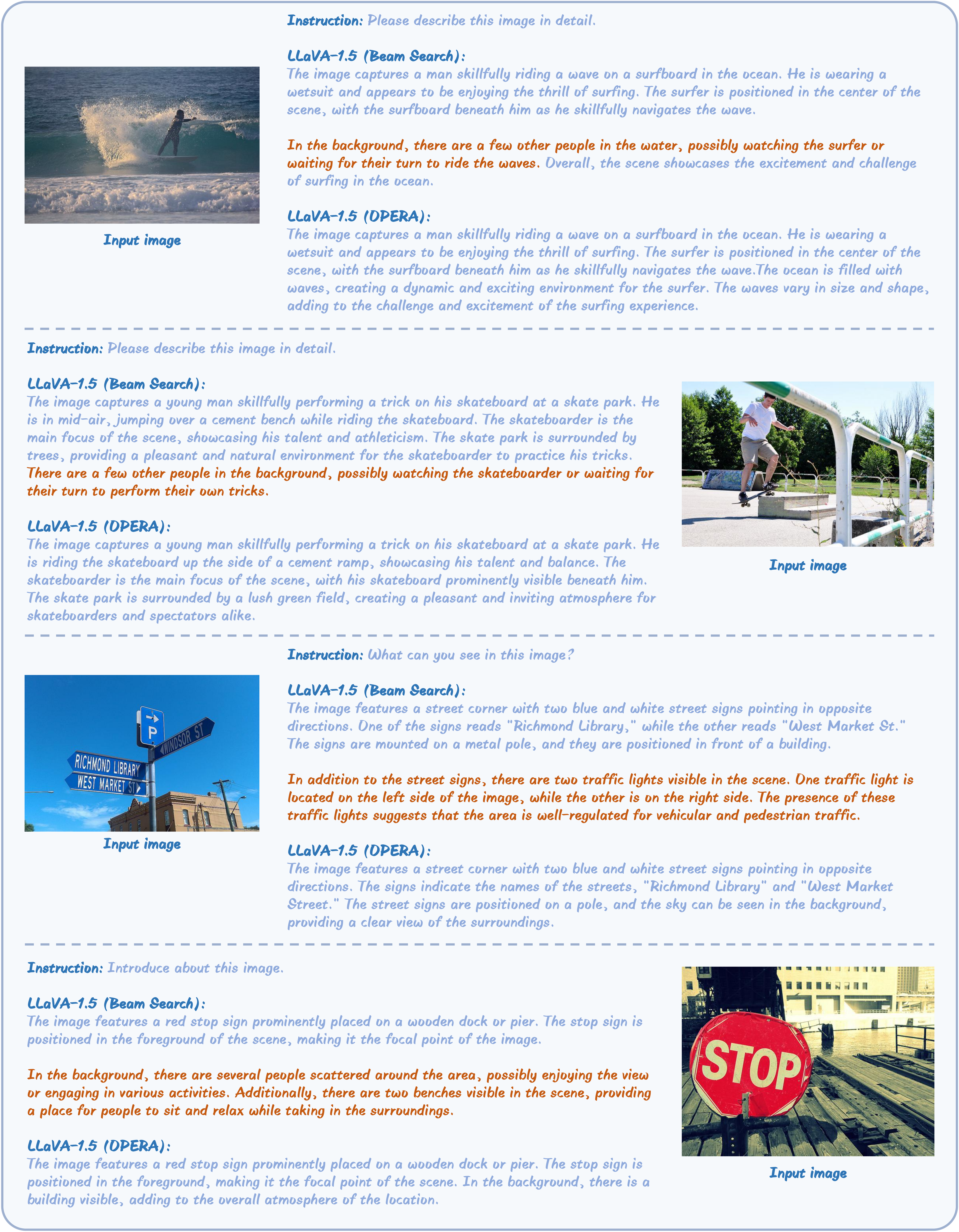}
\caption{OPERA's performance on reducing hallucinations of LLaVA-1.5-7B. }
\label{fig:cases_1}
\end{figure*}

\begin{figure*}[t]
\centering
\includegraphics[width=1.0\linewidth]{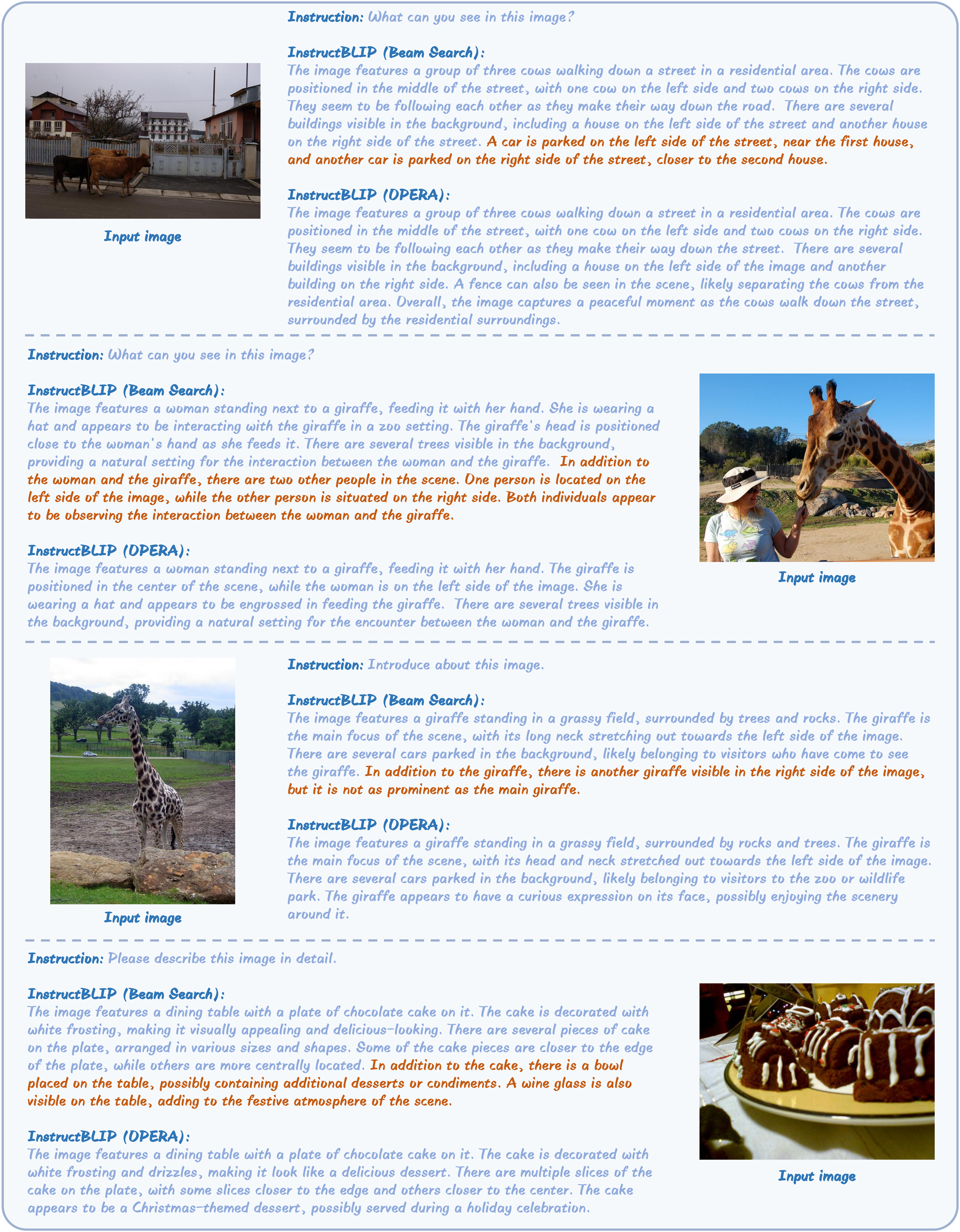}
\caption{OPERA's performance on reducing hallucinations of InstructBLIP-7B. }
\label{fig:cases_2}
\end{figure*}

\begin{figure*}[t]
\centering
\includegraphics[width=1.0\linewidth]{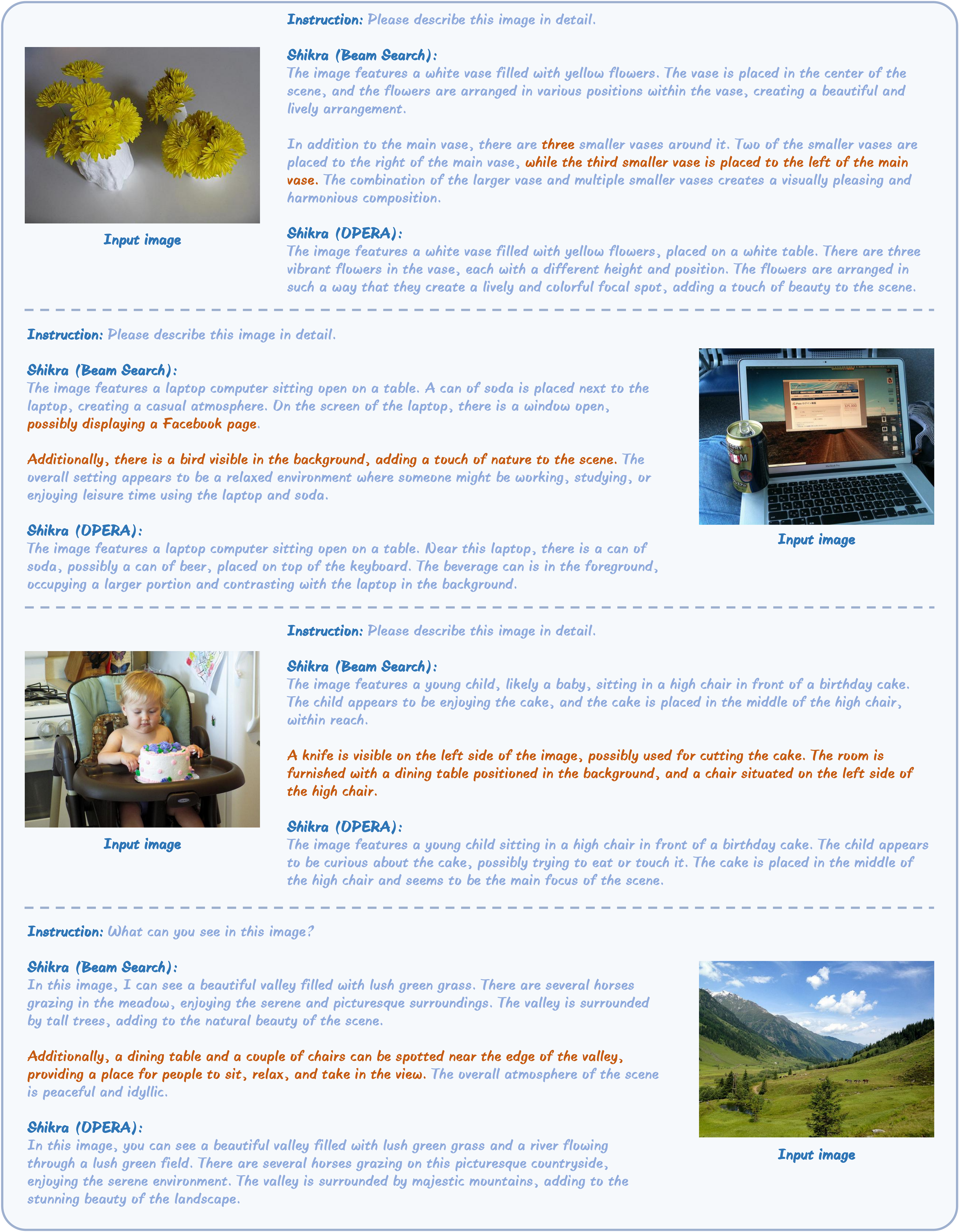}
\caption{OPERA's performance on reducing hallucinations of Shikra-7B. }
\label{fig:cases_3}
\end{figure*}

\subsection{Potentials for Eliminating Repetition}
\label{sec:supp_6}

Repetition is also a problem of MLLMs, usually manifested as the model's incessantly repeating on the particular sentence. 
We notice that OPERA can well handle such repetition, as showcased in \Fref{fig:repetition_case}. 
Interestingly, the self-attention map of repeated sentences appears periodic knowledge aggregation patterns. 
Accordingly, OPERA can help the sequence to retrospect and reallocate at other appropriate vocabularies like ``eos'' token.

\subsection{Qualitative Results}

We provide several cases that proves OPERA's strong ability on mitigating hallucinations. 
These cases uses various MLLMs and different instructions including ``\texttt{Please describe this image in detail.}'', ``\texttt{What can you see in this image?}'', and ``\texttt{Introduce about this image.}''. 
The cases are shown in \Fref{fig:cases}, \Fref{fig:cases_1}, \Fref{fig:cases_2} and \Fref{fig:cases_3} (Please check the next pages).

\section{Limitation \& Social Impact}
\label{sec:supp_1}

In this section, we clarify the weaknesses of our proposed OPERA and the potential social impact incurred by it. 

\noindent\textbf{Limitations.} 
We have identified two main limitations of the proposed approach: 
1) The first limitation lies in it can not address all kinds of the hallucinations of MLLMs. 
It is understandable since our approach serves as a nearly free lunch method for MLLMs without incurring additional costs.
Upon reviewing the failure cases of OPERA, we discern various causes for hallucinated content. 
One likely reason is MLLMs' strong biases in the generated content. 
The knowledge aggregation mechanism of MLLMs causes subsequent token generation to overly rely on summary tokens while neglecting detailed information from the front-most image tokens. 
For instance, MLLMs may easily hallucinate ``cars'' in subsequent tokens when the preceding content mentions ``road''. 
Such hallucinations should blame MLLM's strong bias between ``road'' and ``cars'', which is learned during the training phase. 
In this scenario, OPERA can well handle many cases unless the model's bias is too strong that it is challenging to find a suitable candidate during the retrospection-reallocation phase. 
Another probable reason is that MLLMs' visual perception is not sufficiently robust. MLLMs can be misled by similar shapes, colors of objects, or issues related to low resolution. In these cases, OPERA faces challenges, constrained by the model's visual capabilities.
2) The second limitation is that, OPERA demonstrates marginal gains when addressing hallucinations in short answers ($<$ 10 tokens), primarily due to the hysteresis of knowledge aggregation patterns. 
OPERA excels in handling hallucinations occurring in long sequences. To overcome this limitation, a potential solution is to enhance the metric for detecting knowledge aggregation patterns and increase its sensitivity.

\noindent\textbf{Social impacts.} 
There is no potential for social harm caused by OPERA. 
Instead, it holds the promise to significantly propel the advancement of MLLMs.
OPERA serves as an inspiration for the community to delve into more effective approaches for alleviating MLLMs' hallucination issue without incurring additional costs. 
Such approaches can better generalize on different kinds of MLLMs.

\section{Conclusion}

We introduce OPERA, a novel MLLM decoding method that mitigates hallucination without requiring additional data, knowledge, or training costs. 
It is grounded in an Over-trust Penalty and a Retrospection-Allocation strategy, with the key observation that hallucinations are closely tied to knowledge aggregation patterns in the self-attention matrix, where MLLMs tend to focus on summary tokens, neglecting image tokens and resulting in content hallucination. 
Experiments show our superiority in reducing hallucination on various MLLMs and metrics.

{
    \small
    \bibliographystyle{ieeenat_fullname}
    \bibliography{main}
}


\end{document}